\documentclass[a4paper,fleqn]{cas-dc}

\usepackage[numbers]{natbib}
\usepackage{graphics} 
\usepackage{amsmath} 
\usepackage{amssymb}  
\usepackage{amsfonts}
\usepackage{xcolor}
\usepackage{epsfig}
\usepackage{graphicx}
\usepackage{subfigure}
\usepackage{epstopdf} 
\usepackage{algorithm} 
\usepackage{algpseudocode} 
\usepackage{balance}
\usepackage{stfloats}

\def\tsc#1{\csdef{#1}{\textsc{\lowercase{#1}}\xspace}}
\tsc{WGM}
\tsc{QE}
\tsc{EP}
\tsc{PMS}
\tsc{BEC}
\tsc{DE}

\begin{document}
\let\WriteBookmarks\relax
\def\floatpagepagefraction{1}
\def\textpagefraction{.001}
\shorttitle{Biomimetic Intelligence and Robotics}

\title [mode = title]{Relevant Region Sampling Strategy with Adaptive Heuristic for Asymptotically Optimal Path Planning
}             

\address[1]{Department of Electronic Engineering, The Chinese University of Hong Kong, Shatin N.T., Hong Kong SAR, China}
\address[2]{Department of Electronic and Electrical Engineering, Southern University of Science and Technology, Shenzhen, China}
\address[3]{Shenzhen Research Institute of the Chinese University of Hong Kong, Shenzhen, China}

\author[1]{Chenming Li}[orcid=0000-0001-6322-0834]
\ead{licmjy@link.cuhk.edu.hk}

\author[1]{Fei Meng}
\ead{feimeng@link.cuhk.edu.hk}

\author[1]{Han Ma}
\ead{hanma@link.cuhk.edu.hk}

\author[2]{Jiankun Wang}
\ead{wangjk@sustech.edu.cn}

\author[1,2,3]{Max Q.-H. Meng}
\ead{max.meng@ieee.org}

\cortext[cor2]{This project is supported by Shenzhen Key Laboratory of Robotics Perception and Intelligence (ZDSYS20200810171800001) and the Hong Kong RGC GRF grants \# 14200618 awarded to Max Q.-H. Meng. \textit{(Corresponding authors: Jiankun Wang, Max Q.-H. Meng.)}}

\begin{abstract}
Sampling-based planning algorithm is a powerful tool for solving planning problems in high-dimensional state spaces. 
In this article, we present a novel approach to sampling in the most promising regions, which significantly reduces planning time-consumption.
The RRT\# algorithm defines the Relevant Region based on the cost-to-come provided by the optimal forward-searching tree. 
However, it uses the cumulative cost of a direct connection between the current state and the goal state as the cost-to-go.
To improve the path planning efficiency, we propose a batch sampling method that samples in a refined Relevant Region with a direct sampling strategy, which is defined according to the optimal cost-to-come and the adaptive cost-to-go, taking advantage of various sources of heuristic information.
The proposed sampling approach allows the algorithm to build the search tree in the direction of the most promising area, resulting in a superior initial solution quality and reducing the overall computation time compared to related work.
To validate the effectiveness of our method, we conducted several simulations in both $SE(2)$ and $SE(3)$ state spaces.
And the simulation results demonstrate the superiorities of proposed algorithm.
\end{abstract}


\maketitle

\section{Introduction}

Robot motion planning algorithms are designed to find a trajectory for the robot from its starting point to its destination while avoiding obstacles. 
The robot navigation framework usually simplifies the problem by treating it as a geometric path planning problem to reduce computation complexity. 
While the geometric path planning problem is easier to solve than the motion planning problem, it can still be computationally challenging in some scenarios, such as continuous or high-dimensional spaces.
There are path planning algorithms that use random sampling to solve high-dimensional continuous space planning problems, such as the Rapidly-exploring Random Trees (RRT) \cite{lavalle1998rapidly} and the Probabilistic Roadmap \cite{kavraki1996probabilistic}.
However, these algorithms do not optimize the solution they find. 
To improve the planning efficiency, asymptotically optimal planning algorithms continuously refine the current solution path until it meets the desired quality or the allotted time has expired. 
Although asymptotically optimal planning methods guarantee finding a satisfactory solution path given sufficient computation time, the speed of solving the robot planning problem is crucial for real-time robot operation. 
This speed depends on several key factors, including the speed of finding the initial feasible solution, the quality of the initial solution, and the convergence rate. 
Therefore, it is essential to optimize these factors to ensure efficient path planning for real-time robot operation.

As widely used approaches when solving the planning problems in high-dimensional continuous environments, sampling-based path planning algorithms typically separate the planning process into two stages: sampling and searching \cite{khaksar2016application}.
They samples randomly in the sampling stage, and then use these samples to construct a searching tree or a roadmap in the searching stage. 
Edge evaluation in the searching stage is necessary to ensure that each edge can be executed by the robot.
However, edge evaluation can be expensive and a bottleneck for computation speed in most cases. 
Using inaccurate heuristics or inappropriate strategies can cause the planner to search irrelevant regions, which wastes computational resources and affects real-time performance.
Representative sampling-based path planning methods, such as the RRT \cite{lavalle1998rapidly}, typically use uniform sampling to generate samples and utilize a tree structure to search the space.
The RRT* algorithm \cite{karaman2011sampling}, an asymptotical version of RRT, selects parent vertices based on the cost-to-come value and rewires the tree structure. 
However, RRT* only refines the current tree structure locally.
In contrast, the RRT\# \cite{arslan2013use} proposes to use a replanning function to propagate changes in the relevant part of the graph, rather than rewiring the tree locally.
However, both the RRT* and RRT\# use uniform sampling to produce samples across the entire state space, leading to low sampling efficiency.
Uniform sampling can provide a topology abstraction of the entire state space, however, planners do not need to know information about the entire state space in most planning problems.
The uniform sampling method causes the planner to sample in redundant regions that do not contribute to the solution. 
As a result, this lead to unnecessary edge evaluations and wasted computations.
To address the aforementioned problems, we present a direct sampling technique within a subset of the entire state space, enhancing the efficiency of the sampling stage and allowing the algorithm to concentrate on abstracting the most promising region and optimizing the overall performance.

The cost-to-come and cost-to-go are fairly easy to acquire, and therefore widely used to guide the sampling process.
Conventional methods typically estimate the cost-to-come using non-optimal search trees and estimate the cost-to-go using Euclidean distance to the goal. 
Unfortunately, these estimates are often significantly different from the true cost-to-come and cost-to-go. 
In light of this, we propose utilizing optimal cost-to-come and adaptive cost-to-go estimation methods to guide the sampling process. 
To calculate the optimal cost-to-come, a replanning method is used that propagates changes along the search tree. 
The adaptive cost-to-go estimation employs similar approach as the Adaptively Informed Trees (AIT*) algorithm, which is obtained through a lazy-reverse tree technique.
As a result of the optimal cost-to-come and the adaptive cost-to-go estimation, our method focuses on taking samples from the areas that have the highest probability of improving the solution paths, which is closely relevant to the current query and current tree. 
Therefore, we have named this sampling technique the Relevant Region sampling strategy.


\begin{figure*}[ht]
    \centering
        \begin{minipage}[t]{1\linewidth}
            \subfigure[]{
                \begin{minipage}[t]{0.24\linewidth}
                    \centering
                    \includegraphics[width=0.9\linewidth]{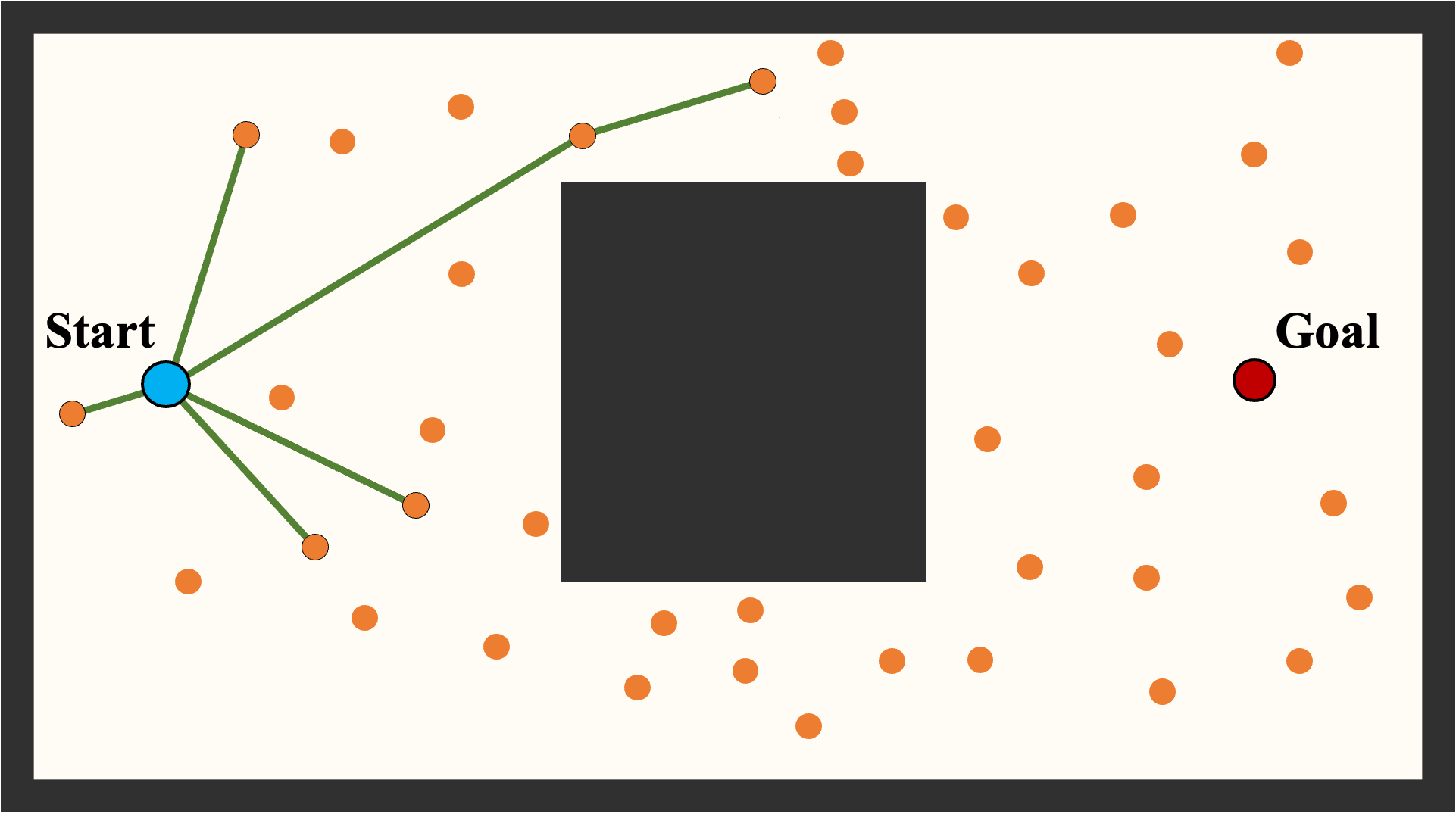}
                \end{minipage}%
            }
            \subfigure[]{
                \begin{minipage}[t]{0.24\linewidth}
                    \centering
                    \includegraphics[width=0.9\linewidth]{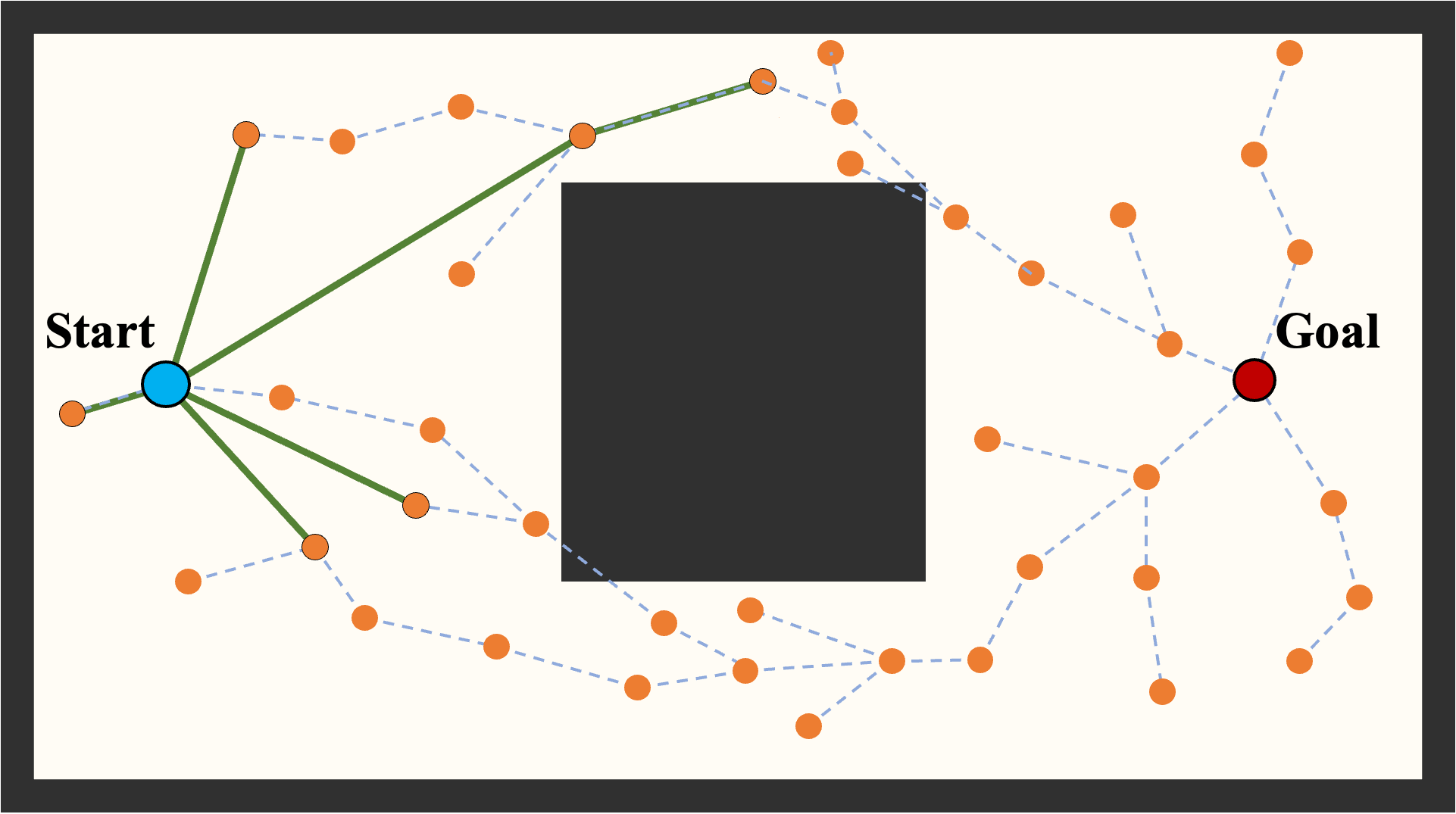}
                    \label{collisionFigB}
                \end{minipage}%
            }
            \subfigure[]{
                \begin{minipage}[t]{0.24\linewidth}
                    \centering
                    \includegraphics[width=0.9\linewidth]{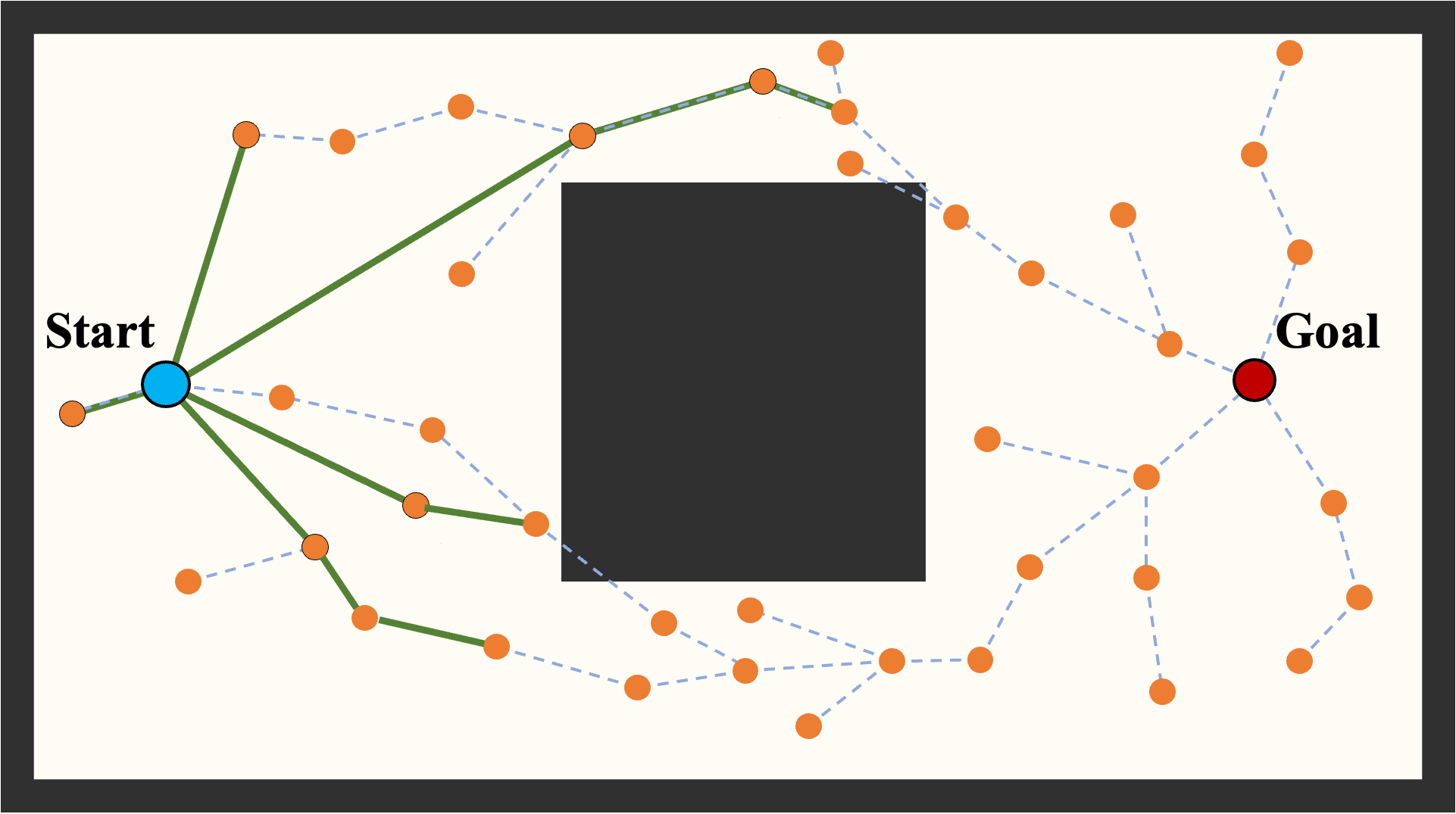}
                    \label{collisionFigC}
                \end{minipage}
            }%
            \subfigure[]{
                \begin{minipage}[t]{0.24\linewidth}
                    \centering
                    \includegraphics[width=0.9\linewidth]{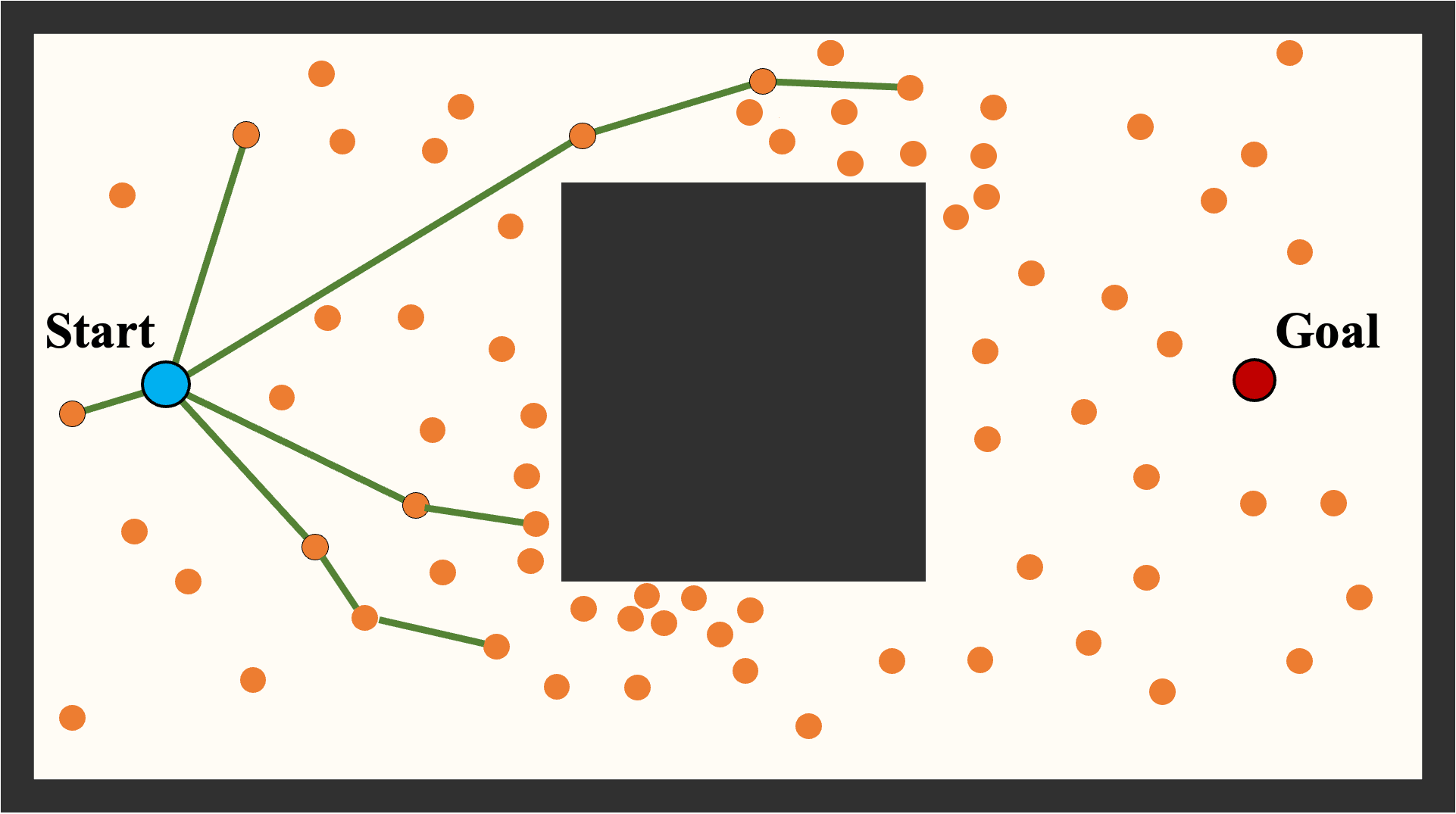}
                \end{minipage}
            }%
        \end{minipage}%
    \centering
    \caption{ The schematic for the proposed method in a simple 2D environment, where the black blocks, the orange point, solid green lines, and the dashed grey lines demonstrate the obstacles, the sampling points, edges in the forward-searching tree, and edges in the reverse-searching trees, respectively.
              Fig. (a) shows that our method sample a batch of points in the current iteration.
              Fig. (b) shows the lazy reverse-searching stage.
              Fig. (c) shows the forward-searching stage.
              And Fig. (d) shows the current lazy reverse-searching tree is discarded, and a new batch of samples is added in the next iteration, where the newly added samples are more likely to be concentrated in areas that are `difficult to navigate'.
              }
    \label{schematicIllustration}
\end{figure*}

To describe the workflow of our method, we use a simple path planning schematic in a 2D environment illustrate it.
This diagram makes it easy to visualize the process, as shown in Fig. \ref{schematicIllustration}.
In the very first, the planner draws random samples in the free space and builds a lazy-reverse tree to provide a problem-specific cost-to-go estimation. 
Based on this estimation, the planner builds a forward search tree to explore the space in terms of the cost-to-go estimation.
When the forward search tree is updated, a replanning technique propagates the update to find a better global connection and the optimal cost-to-come for the newly added vertex. 
After each batch of samples is fully explored, we take a new batch of samples based on the optimal cost-to-come and the last cost-to-go estimation.
Then a new lazy-reverse tree is constructed to guide the searching of the forward tree, until a termination condition is met.

Simulations are all carried through the benchmark platform of the Open Motion Planning Library (OMPL) \cite{sucan2012open} \cite{moll2015benchmarking}. 
To demonstrate the generality of the proposed method, we evaluate it in both $SE(2)$ and $SE(3)$ simulation environments.
We show that by combining forward tree replanning with the proposed sampling technique, the planning performance is improved.

In this article, we introduce the related work in Section II. 
The problem definition is described in Section III, and the problem-specific heuristic is discussed as well.
Then, we present the mathematical details and the proposed algorithm in Section IV. 
Simulations are carried out to validate the effectiveness of the proposed method, and the results are shown in Section V.  
Finally, we draw our conclusions in Section VI.

\section{Related Work}

\subsection{Sampling-based Motion Planning Method}

Plenty of modifications are proposed to enhance the performance of the RRT algorithm \cite{lavalle1998rapidly} such as the RRT* algorithm \cite{karaman2011sampling}.
The rewiring stage of the RRT* only rewires locally, which means the global optimization of the current tree is ignored. 
The RRT\# \cite{arslan2013use} proposes to find the global optimality in each rewiring stage with dynamic programming.
Dynamic programming is also used in the Fast Marching Tree (FMT*) method \cite{janson2015fast} to grow the searching tree.
The Informed sampling strategy \cite{gammell2014informed, gammell2018informed} is proposed to overcome the drawback of uniform sampling.
Using the neural network to reinforce the sampling stage to enhance the sampling efficiency \cite{wang2020neural, li2021efficient, qureshi2019motion} is proved as a promising technique.



\subsection{Batch Sampling Technique}

The FMT* \cite{janson2015fast} introduces the thought of batch sampling into the robot motion planning field.
The FMT* samples a batch of points and constructs the searching tree according to this batch of samples.
The Batch Informed Trees (BIT*) \cite{gammell2015batch, gammell2020batch} method is developed based on the Informed RRT*, besides, the BIT* absorbs the thoughts in the FMT* method \cite{janson2015fast} and the Lifelong Planning A* (LPA*) algorithm \cite{koenig2004lifelong}.
The Regionally Accelerated Batch Informed Trees (RABIT*) \cite{choudhury2016regionally} aims to solve the difficult-to-sample planning problem, like the narrow passage problem.
The Fast-BIT* \cite{holston2017fast} modifies the edge queue and searches the initial solution more aggressively. 
The Greedy BIT* \cite{chen2021greedy} uses the greedy searching method to generate the initial solution faster and accelerate the convergence speed.
But these greedy-based methods often fail to assist the searching procedure without an accurate heuristic estimation method. 
The Adaptively Informed Trees (AIT*) \cite{strub2020adaptively} and the Advanced BIT* (ABIT*) \cite{strub2020advanced} proposed by Strub and Gammell are developed based on the BIT* as well. 
The AIT* calculate a relatively accurate heuristic estimation with a lazy reverse-searching tree.
The ABIT* proposes to utilize inflation and truncation to balance the exploitation and exploration in the increasingly complex Random Geometric Graph (RGG) \cite{penrose2003random}.
Though the AIT* and the ABIT* achieve significant improvements, their sampling regions are not compact enough, and the sampling efficiency will be critically low in the complex environment.

\subsection{Relevant Region Sampling Strategy}

The concept of 'relevant' is first proposed in the searching-based robot path planning method like the A* \cite{hart1968formal}. 
In the A* algorithm, the set of expanded vertices is relevant to the query, such that the A* algorithm could expand a smaller set of vertices than the Dijkstra's algorithm \cite{dijkstra1959note}.  
The concept of the Relevant Region is formally introduced in \cite{arslan2013use}, whose sum of the optimal cost-to-come and cost-to-go heuristic is less than the cost of the current optimal solution. 
Since the Relevant Region is the most promising area for improving the solution, a straightforward modification would be reducing the likelihood of sampling outside of it.
Three different metrics are used to achieve this in \cite{arslan2015dynamic}, the modified versions achieve better performance in the convergence speed than the RRT\# method.
The methods described in \cite{arslan2013use} and \cite{arslan2015dynamic} use the rejection method for sampling, which is not efficient since the Relevant Region is a small subset of the whole state space in most scenarios.
The direct sampling method is illustrated to overcome this drawback, and the details are described in \cite{joshi2020relevant}.
However, they all use the cumulative cost along the direct connection between the current state and the goal state as the cost-to-go. 
This approach results in inaccurate estimated cost-to-go in most scenarios.


\subsection{Bi-directional Searching Method}

The RRT and RRT* methods may not always discover a solution within the allotted time, particularly when dealing with narrow passages
The RRT-Connect \cite{kuffner2000rrt} is proposed to find the initial solution faster. 
However, the approach described in \cite{kuffner2000rrt} is not asymptotically optimal. Therefore, its successor, RRT-Connect, is also not asymptotically optimal. 
To overcome this, an enhanced version of the bidirectional searching RRT is introduced in \cite{klemm2015rrt} to guarantee the asymptotical optimality.
To take advantage of the benefits of bi-directional search, the kinematic constraints are taken into consideration in the bi-directional search method to generate executable trajectories efficiently \cite{wang2021kinematic}.


One drawback of the Informed RRT* \cite{gammell2014informed} \cite{gammell2018informed} is that it uses the RRT* to search the whole state space before finding the initial solution.
Therefore, the Informed RRT* often fails to find the solution in the required period, same as the RRT*.
By combining the advantages of both the Informed and the RRT*-Connect, the Informed RRT*-Connect \cite{2020Informed} proposes to use the RRT*-Connect to generate the initial solution and use the Informed sampling strategy to constrain the sampling region after the initial solution is found.
Besides, the AIT* \cite{strub2020adaptively} can also be viewed as a bi-directional searching method.

\section{Problem Definition}

Consider the state space $\mathcal{X}$, which is the subset of $\Re^d$.
$\Re^d$ is the whole $d$-dimensional space, and $d$ is a positive integer.
$\mathcal{X}_{obs}$ shows the space occupied by the obstacles, the free space is defined as $\mathcal{X}_{free} = \mathcal{X} \setminus \mathcal{X}_{obs} $.
The $x \in \mathcal{X}$ represents any state in the state space.
The source point $x_{start}$ is the initial state of the robot.
The destination is a region represented by $\mathcal{X}_{goal}$.
The source point and the destination in a valid planning problem must be defined within the free space $x_{start} \in \mathcal{X}_{free} \ \& \ \mathcal{X}_{goal} \subseteq \mathcal{X}_{free}$.
The motion planning problem is defined as:


\begin{equation}
\begin{aligned}
& \pi \in [0, 1] \to \mathcal{X}_{free}, \\  
& s.t. \ \pi(0) = x_{start}, \ \pi(1) \in \mathcal{X}_{goal}, \ 
\pi(s) \in \mathcal{X}_{free},
\forall s \in [0, 1] .
\label{MotionPlanningDefinition}
\end{aligned}
\end{equation}

The optimization objective can be minimizing the trajectory length, maximizing the minimum clearance, or any objects that could be mathematically defined. 
It can also be set as the sum of individual optimization objectives and form a hybrid optimization objective.
For simplicity, the optimization objective in the proposed method is set as minimizing the trajectory length.
Note that it could be extended to meet the specific planning requirement.
The $v \in \mathcal{T}$ denotes any vertex in the tree.
Assume there are two vertices $v_1, \ v_2 \in \mathcal{T}$, where $v_2$ is the descendant of $v_1$.
The cost between any two vertices $v_1, \ v_2$ is calculated with the intergal cost along the tree from $v_1$ to $v_2$, denotes as $d_{\mathcal{T}}(v_1, \ v_2)$.
Let $\Pi$ denote the set of all feasible solutions.
With the definition of trajectory cost, the optimization objective in our method is written as:

\begin{equation}
    \begin{aligned}
    & \pi* = \operatorname*{\arg \min}_{\pi \in \Pi} d_{\mathcal{T}}(v_{start}, v_{goal}) \\
    & s.t. \ \pi(0) = v_{start}, \ \pi(1) \in \mathcal{X}_{goal}, \pi(s) \in \mathcal{X}_{free},  \forall s \in [0, 1] .
\label{TrajectoryLengthOptimizationObjective}
\end{aligned}
\end{equation}

The cost estimation between any two states $x_1, x_2 \in \mathcal{X}$ can be set as the cumulative cost along the direct connection. 
The cost for the unit distance is usually deemed as $1$, the heuristic can be calculated with the Euclidean metric: $ h(x_1, x_2) = \left\| x_2 - x_1 \right\|_2$.
However, using this metric as the heuristic estimation method between any two states is not promising.
The direct connection is highly likely to collide with the obstacle, which can mislead the searching procedure, especially in complex environments.

The RRT algorithm contains two stages: the sampling and the searching stage. The sampling stage can be viewed as the abstracting process of the state space $\mathcal{X}$, and the searching stage will construct the searching tree $\mathcal{T}$ based on this abstracted state space. 
These two phases are performed alternatively in each iteration. 
We use the $c_{cur}$ to represent the cost of the current optimal solution. 
Before finding the initial feasible solution, the $c_{cur}$ is set to an infinitely large number $+\infty$.

\subsection{Adaptive Heuristic Estimation}

The adaptive heuristic estimation method was first proposed by the AIT* algorithm \cite{strub2020adaptively}, which is able to provide a problem-specific heuristic without overburdening computation resources.
We borrowed this idea to provide a fairly accurate estimation of the cost-to-go, and used it to guide the algorithm to take samples in relevant regions.

Since calculating the heuristic along the direct connection is inaccurate, the planner uses a lazy reverse-searching tree $\mathcal{T}_{\mathcal{R}}$ to provide the cost-to-go estimation $h_{\mathcal{T}_{\mathcal{R}}}(x)$ of any state $x$ in the state space, which is analogous to the AIT* algorithm \cite{strub2020adaptively}.
The $\mathcal{T}_{\mathcal{R}}$ is constructed without edge evaluation because edge evaluation is the most time-consuming procedure in the majority of motion planning scenarios.
To construct the $\mathcal{T}_{\mathcal{R}}$, the planner separates the sampling stage and searching stage into two separate modules instead of performing them alternatively in each iteration.
In the sampling stage, the planner uses our sampling strategy to generate a batch of sampling points in $\mathcal{X}_{free}$. 
Then, in the searching stage, the planner constructs the $\mathcal{T}_{\mathcal{R}}$ and $\mathcal{T}_{\mathcal{F}}$ in terms of the current RGG.
Since the $\mathcal{T}_{\mathcal{R}}$ is constructed without edge evaluation, the edges in the $\mathcal{T}_{\mathcal{R}}$ may collide with the obstacles or not satisfy the constraints.
To deal with this, the $\mathcal{T}_{\mathcal{R}}$ will be updated incrementally upon collision, besides, AIT* uses a black and white list approach to prevent re-checking of the collided connections.
In our proposed method, since we do not need edges in the $\mathcal{T}_{\mathcal{R}}$ to be executable, we borrow this idea to use the $\mathcal{T}_{\mathcal{R}}$ to provide a relatively accurate cost-to-go estimation and guidance for sampling.

\section{Methodology}


\subsection{Relevant Region}





In this article, we use the optimal forward-searching tree $\mathcal{T}_{\mathcal{F}}$ to provide the cost-to-come estimation $h_{\mathcal{T}_{\mathcal{F}}}(v)$ of any vertex $v \in \mathcal{T}_{\mathcal{F}}$.
With the optimization objective, the $h_{\mathcal{T}_{\mathcal{F}}}(v) = d_{\mathcal{T}_{\mathcal{F}}}(v_{start}, \ v)$ is defined as the cumulative cost from the initial state $x_{start}=v_{start} \in \mathcal{T}_{\mathcal{F}}$ to the vertex $v$ via the current tree $\mathcal{T}_{\mathcal{F}}$.
And the planner guarantees that the newly added vertex is optimally connected to $\mathcal{T}_{\mathcal{F}}$ with a global replanning function.

The Relevant Region is defined as a subset of $\mathcal{X}_{free}$ of which the cardinality is smaller than the Informed sampling set \cite{gammell2014informed}. 
Our Relevant Region sampling strategy uses the optimal forward-searching tree $\mathcal{T}_{\mathcal{F}}$ to provide the optimal cost-to-come estimation.
And using the lazy reverse-searching tree $\mathcal{T}_{\mathcal{R}}$ to estimate the cost-to-go of the state $x$ in the RGG, the estimated result can be represented as $h_{\mathcal{T}_{\mathcal{R}}}(x) = d_{\mathcal{T}_{\mathcal{R}}}(x, \ x_{goal})$.
With the $h_{\mathcal{T}_{\mathcal{F}}}$ and $h_{\mathcal{T}_{\mathcal{R}}}$, the definition of relevant vertex is shown in (\ref{RelevantVertices}), where the $\mathcal{V}$ is the vertices set of the optimal forward-searching tree $\mathcal{T}_{\mathcal{F}}$ and $v \in \mathcal{V}$ is any vertex belongs to the $\mathcal{T}_{\mathcal{F}}$.


\begin{equation}
\begin{aligned}
\mathcal{V}_{rel} = \{ v \in \mathcal{V} \ | \ h_{\mathcal{T}_{\mathcal{F}}}(v) + h_{\mathcal{T}_{\mathcal{R}}}(v) < c_{cur} \}.
\label{RelevantVertices}
\end{aligned}
\end{equation}

We define the ball centered at the vertex $v$ with radius $\epsilon$ as:

\begin{equation}
\begin{aligned}
\mathcal{B}^{\epsilon}(v) = \{ x \in \mathcal{X} \ | \ \left\| x - v \right\|_2 < \epsilon, v \in \mathcal{V}_{rel}\}.
\label{RelevantBall}
\end{aligned}
\end{equation}

With the $h_{\mathcal{T}_{\mathcal{F}}}(v)$ and the $h_{\mathcal{T}_{\mathcal{R}}}(x)$, the estimated solution cost that pass by the state $x \in \mathcal{B}^{\epsilon}(v)$ can be written as $\hat{f}_v(x) = h_{\mathcal{T}_{\mathcal{F}}}(v) + d(v, x) + h_{\mathcal{T}_{\mathcal{R}}}(x)$, where the $d(v, x)$ denotes cost from vertex $v$ to state $x$.
The Relevant Region with the adaptive heuristic estimation of the current forward-searching tree is defined as (\ref{RelevantRegion}).

\begin{equation}
\begin{aligned}
\mathcal{X}^{\epsilon}_{rel} = \{x \ | \ x \in \mathcal{B}^{\epsilon}(v), \hat{f}_v(x) < c_{cur} \}.
\label{RelevantRegion}
\end{aligned}
\end{equation}

\subsection{Relevant Region Sampling Strategy}

The ultimate optimization objective is shown in (\ref{TrajectoryLengthOptimizationObjective}).
To meet the optimization requirement with fewer sampling points, we propose a novel method to generate samples in the most promising region.
Let the new sample be $x_{new} \in \mathcal{X}_{free}$, and it is related to the relevant point $v_{rel} \in \mathcal{V}_{rel}$, where $v_{rel}$ is chosen from the queue $\mathcal{Q}$ with the highest priority.
The $\mathcal{Q}$ is defined in terms of (\ref{WeightFunction}). 
The relationship between $x_{new}$ and $v_{rel}$ is $x_{new} = v_{rel} + \gamma e$.
where $\{ e\in E \ | \ e \in \Re^d, \left\| e \right\|_2 = 1\}$ denotes the direction for expanding which points from $v_{rel}$ to $x_{new}$, and $\gamma$ is a positive number which indicats the maximum promising cost magnitude to travel along the direction $e$, bounded by $\left(0, \gamma_{max} \right]$.
The sampling direction $e$ is generated by utilizing the information provided by the $\mathcal{T}_{\mathcal{R}}$, which is the direction along the edge of the $\mathcal{T}_{\mathcal{R}}$ pointing to the goal. 
Then we will use the solution of the following optimization problem to guide the sampling.

\begin{equation}
\begin{aligned}
\max & \ \gamma, \\
subject \ to: & \ \hat{f}_v(x) < c_{cur}, \\
& \ \gamma \in (0, \gamma_{max}].
\label{OptimizationProblem}
\end{aligned}
\end{equation}

The inequality in (\ref{OptimizationProblem}) equals to $h_{\mathcal{T}_{\mathcal{F}}}(v) + d(v, x) + h_{\mathcal{T}_{\mathcal{R}}}(x) < c_{cur}$. Then we will have:

\begin{equation}
\begin{aligned}
h_{\mathcal{T}_{\mathcal{R}}}(x) < c_{cur} - h_{\mathcal{T}_{\mathcal{F}}}(v) - \gamma.
\label{Reformate_1}
\end{aligned}
\end{equation}

Note that the right-hand side of the (\ref{Reformate_1}) is always positive,
we can obtain another inequality $\gamma < c_{cur} - h_{\mathcal{T}_{\mathcal{F}}}(v)$.
If the right-hand side $c_{cur} - h_{\mathcal{T}_{\mathcal{F}}}(v)$ is smaller than $0$, the outgoing edges connecting to $v$ and its children will be erased from the forward-searching tree $\mathcal{T}_{\mathcal{F}}$ iteratively.

By moving the $d(v, x) = \gamma$ in (\ref{OptimizationProblem}) to the left-hand side, we can get:

\begin{equation}
\begin{aligned}
\gamma < c_{cur} - h_{\mathcal{T}_{\mathcal{F}}}(v) - h_{\mathcal{T}_{\mathcal{R}}}(x).
\label{OptimizationSolution}
\end{aligned}
\end{equation}

So the (\ref{OptimizationSolution}) is the solution for the best cost magnitude to travel along direction $e$.
After giving the $\gamma$ an upper bound $\gamma_{max}$ for expanding, the final solution is:

\begin{equation}
\begin{aligned}
\gamma_{rel} =
min(c_{cur} - h_{\mathcal{T}_{\mathcal{F}}}(v) - h_{\mathcal{T}_{\mathcal{R}}}(x),\ \gamma_{max} ).
\label{GammaRelevant}
\end{aligned}
\end{equation}



The calculated step size $\gamma_{rel}$ may be a negative number since the $h_{\mathcal{T}_{\mathcal{F}}}(v)$ and $h_{\mathcal{T}_{\mathcal{R}}}(x)$ always overestimate the cost.
In that case, the result will be invalid.

\begin{figure}[t]
    \centering
    \includegraphics[width=0.5\textwidth]{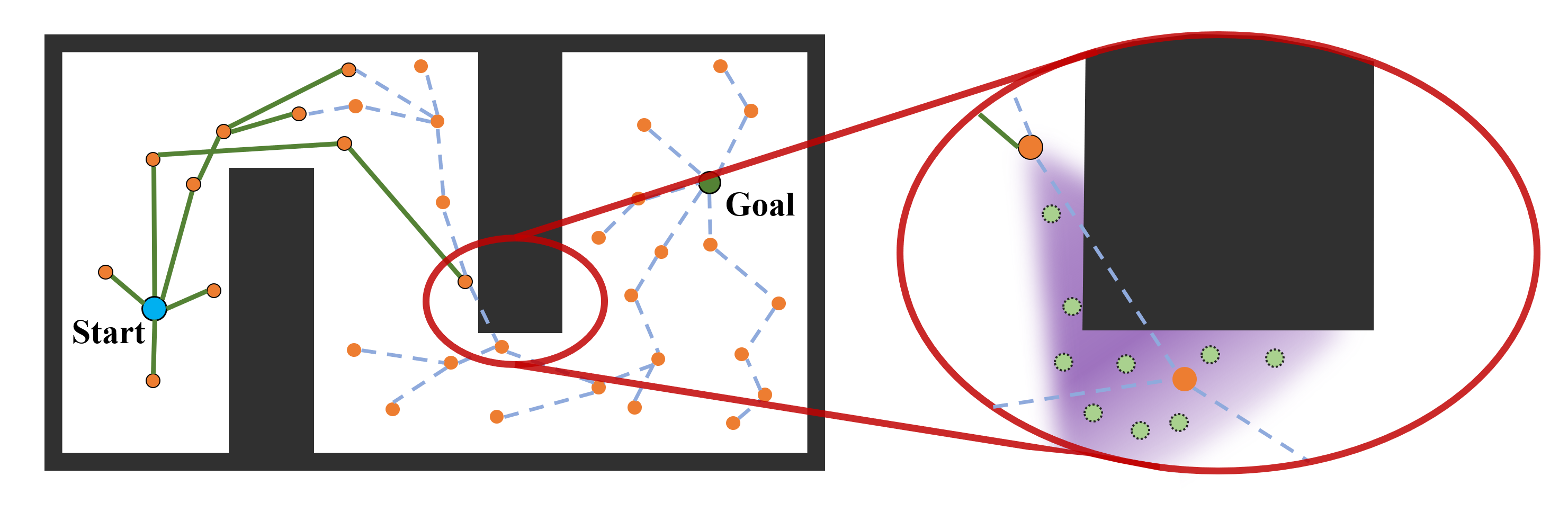}
    \caption{We add randomness to the $e$ and $\gamma$ since the lazy reverse-searching method can not guarantee edges in the $\mathcal{T}_{\mathcal{R}}$ are valid. 
    In this figure, the black blocks, solid green lines, dashed blue lines, and orange points represent the obstacles, forward tree, reverse tree, and the points in the current RGG, respectively.
    The right part of the figure is the local zoom of the left part.
    The purple circular sector shows the region where we may take samples in terms of the vertex $v_{rel}$.
    The region with darker purple has a higher probability of being sampled.}
    \label{RandomDirection}
\end{figure}

\begin{algorithm}[t]
    \caption{Relevant Sampling Strategy with Adaptive Heuristic Estimation for Asymptotically Optimal Motion Planning}
    \label{MainFunction}
        \hspace*{\algorithmicindent} \textbf{Input: $x_{start} \in \mathcal{X}_{free}$, $\mathcal{X}_{goal} \subset \mathcal{X}_{free}$ }; \\
        \hspace*{\algorithmicindent} \textbf{Output: $\mathcal{T}_{\mathcal{F}}$};
        \begin{algorithmic}[1]
        \State Init($\mathcal{T}_{\mathcal{F}}$, $\mathcal{T}_{\mathcal{R}}$, $\mathcal{V}_{sol}$);
        \While{\textbf{not} TerminateCondition()}
            \State $\mathcal{X}_{\mathcal{S}_{reuse}} \gets$ Prune($\mathcal{X}_{\mathcal{S}}$, $\mathcal{T}_{\mathcal{F}}$, $\mathcal{T}_{\mathcal{R}}$);   \label{pruneFunction}
            \State $\mathcal{X}_{\mathcal{S}_{new}} \gets$ Sample($m$, $P_{Inf}$, $\mathcal{T}_{\mathcal{F}}$);  \label{SampleLine}
            \State $\mathcal{X}_{\mathcal{S}} \gets \mathcal{X}_{\mathcal{S}_{reuse}} \cup \mathcal{X}_{\mathcal{S}_{new}}$;
            \State $\mathcal{T}_{\mathcal{R}}$ $\gets$ BuildReverseTree($\mathcal{X}_{\mathcal{S}}$);
            \State $\mathcal{T}_{\mathcal{F}}$ $\gets$ BuildForwardTree($\mathcal{T}_{\mathcal{F}}$, $\mathcal{T}_{\mathcal{R}}$); \label{BuildForwardTreeLine}
        \EndWhile
        \State \textbf{return} $\mathcal{T}_{\mathcal{F}}$;
    \end{algorithmic}
\end{algorithm}





\begin{algorithm}[t]
	\caption{Sample($m$, $P_{Inf}$, $\mathcal{T}_{\mathcal{F}}$)}
    \label{SampleFunction}
        \begin{algorithmic}[1]
            \State $\mathcal{X}_{\mathcal{S}_{new}} \gets \emptyset $;
            \State $\mathcal{Q}$ = WeightVertices($\mathcal{T}_{\mathcal{F}}$); \label{WeightVertices}
            \State $\mathcal{X}_{\mathcal{S}_{new}} \gets $ $\mathcal{X}_{\mathcal{S}_{new}} \cup $ SampleRel($Q$); \label{RelevantSampleLine}
            \State $\mathcal{X}_{\mathcal{S}_{new}} \gets $ $\mathcal{X}_{\mathcal{S}_{new}} \cup $ SampleInf($m - \mathcal{X}_{\mathcal{S}_{new}}.size()$);
            \State \textbf{return} $\mathcal{X}_{\mathcal{S}_{new}}$;
        \end{algorithmic}
\end{algorithm}

\begin{equation}
    \begin{aligned}
        \label{WeightFunction}
        \mathcal{W} = \lambda_1 n_{s} + \lambda_2 n_{o} + \lambda_3 \frac{h_{\mathcal{T}_{\mathcal{F}}}(v) + h_{\mathcal{T}_{\mathcal{R}}}(v)}{c_{cur}}.
    \end{aligned}
\end{equation}

\begin{algorithm}[t]
	\caption{SampleRel($\mathcal{Q}$)}
    \label{SampleRelevantRegion}
        \begin{algorithmic}[1]
            \State $ExpandTimes$ = $int( (1 - P_{Inf}) m / \mathcal{Q}.size() )$;    \label{NumOfReleventQueue}
            \For{$v_{elem}$ \textbf{in} $\mathcal{Q}$}
                \State $\gamma$, $e$ = EstimateStepSizeAndDirection($v_{elem}$);
                \For{$iter$ \textbf{in} $iter = 1, 2, 3, \dots, ExpandTimes$}
                    \State $\hat{\gamma}$, $\hat{e}$ = GaussianNoise($\gamma$, $e$);
                    \State $\mathcal{X}_{\mathcal{S}_{new}}.append(v_{elem} + \hat{\gamma}\hat{e})$;
                \EndFor
            \EndFor
            \State \textbf{return} $\mathcal{X}_{\mathcal{S}_{new}}$;
        \end{algorithmic}
\end{algorithm}

\begin{algorithm}[t]
    \caption{BuildForwardTree($\mathcal{T}_{\mathcal{F}}$, $\mathcal{T}_{\mathcal{R}}$)}
    \label{BuildForwardTreeFunction}
        \begin{algorithmic}[1]
            \While{$\mathcal{Q}_{V}$.best() $\leq$ $\mathcal{Q}_{E}$.best()} \label{ExpandNextVertex}
            \State ExpandNextVertex($\mathcal{Q}_{V}$, $\mathcal{Q}_{E}$, $c_{cur}$);    
            \EndWhile
            \While{True}
            \State ($v_{min}$, $x_{min}$) $\gets$ PopBest($\mathcal{Q}_{E}$);
            \If{$g_{\mathcal{T}}(v_{min}) + \hat{c}(v_{min}, x_{min}) + \hat{h}(x_{min}) < c_{cur}$}
                \If{$g_{\mathcal{T}}(v_{min}) + \hat{c}(v_{min}, x_{min}) < g_{\mathcal{T}}(x_{min})$}
                    \State $c_{edge} \gets c(v_{min}, x_{min})$;
                    \If{$g_{\mathcal{T}}(v_{min}) + c_{edge} + \hat{h}(x_{min}) < c_{cur}$}
                        \If{$g_{\mathcal{T}}(v_{min}) + c_{edge} < g_{\mathcal{T}}(x_{min})$}   \label{EndPostpone}
                            \State $\mathcal{T}_{\mathcal{F}} \gets$ Extend($\mathcal{T}_{\mathcal{F}}$, $x_{min}$);
                            \State $\mathcal{T}_{\mathcal{F}}$ $\gets$ Replan($\mathcal{T}_{\mathcal{F}}$);
                            \State $c_i \gets min_{v_{goal} \in \mathcal{V}_{sol} }\{ g_\mathcal{T}(v_{goal}) \}$;
                        \EndIf
                    \EndIf
                \EndIf
            \Else 
                \State $\mathcal{Q}_V \gets \emptyset$; $\mathcal{Q}_E \gets \emptyset$;
                \State \textbf{break};
            \EndIf
            \EndWhile
        \end{algorithmic}
\end{algorithm}

In the implementation, each vertex will be used to calculate random sampling points multiple times with different random distortions in each iteration.
We add the random distortions to both the $e$ and $\gamma$ because edges in the $\mathcal{T}_{\mathcal{R}}$ are not valid under the constraints, and direct search along $e$ is not promising.
The searching direction and the expanding distance with random distortion are denoted as $\hat{e}$ and $\hat{\gamma}$, respectively, and the new sample is $\hat{x}_{new} = v_{rel} + \hat{\gamma}\hat{e}$.
We illustrate this in Fig. \ref{RandomDirection}.
The pale green points circled by the dotted lines in the right-hand side of Fig. \ref{RandomDirection} indicate the newly added sampling points with our direct sampling method in the current batch. 

\subsection{Proposed Algorithm}

\begin{figure*}[t]
    \centering
        \begin{minipage}[t]{1\linewidth}
            \subfigure[]{
                \begin{minipage}[t]{0.235\linewidth}
                    \centering
                    \includegraphics[width=0.9\linewidth]{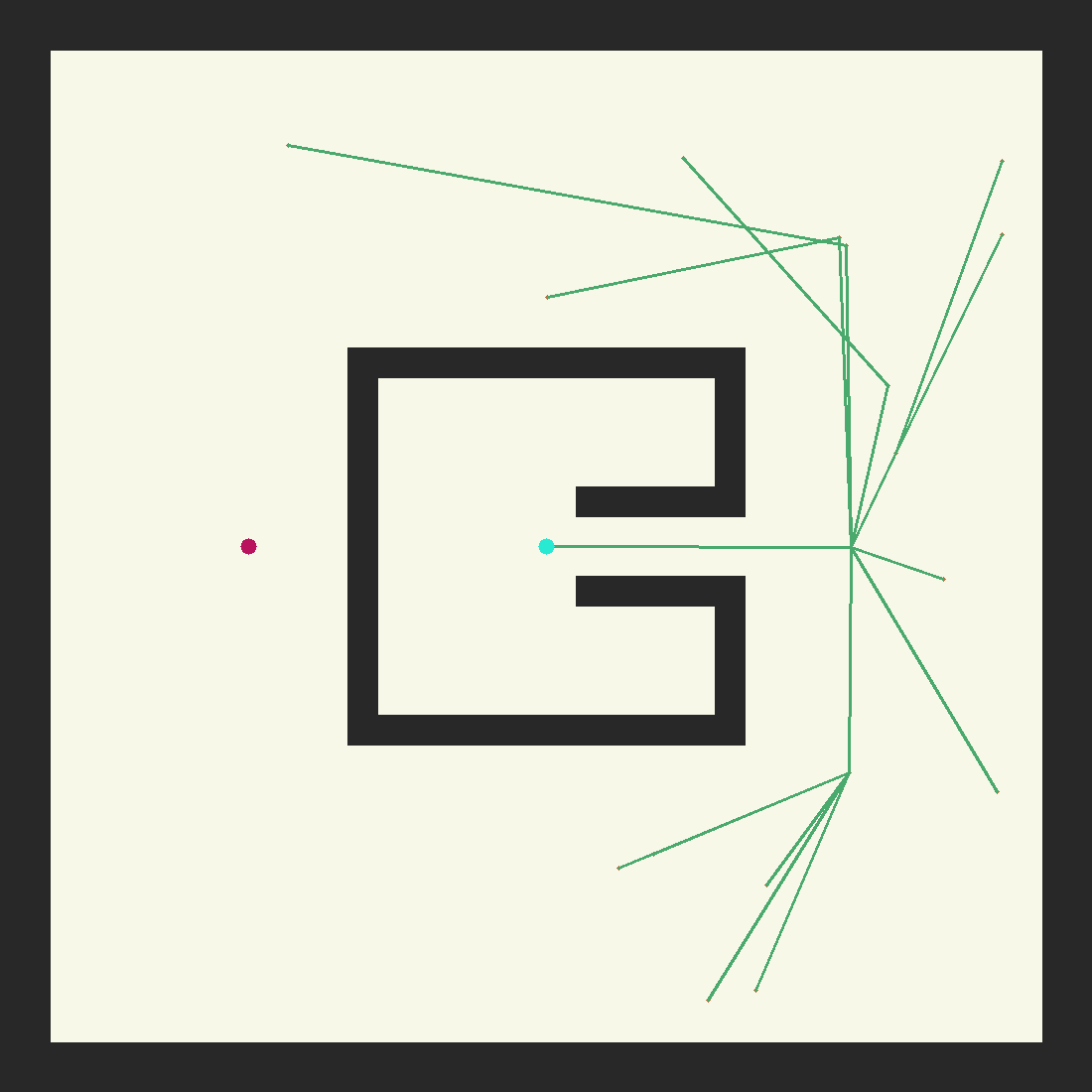}
                \end{minipage}%
            }
            \subfigure[]{
                \begin{minipage}[t]{0.235\linewidth}
                    \centering
                    \includegraphics[width=0.9\linewidth]{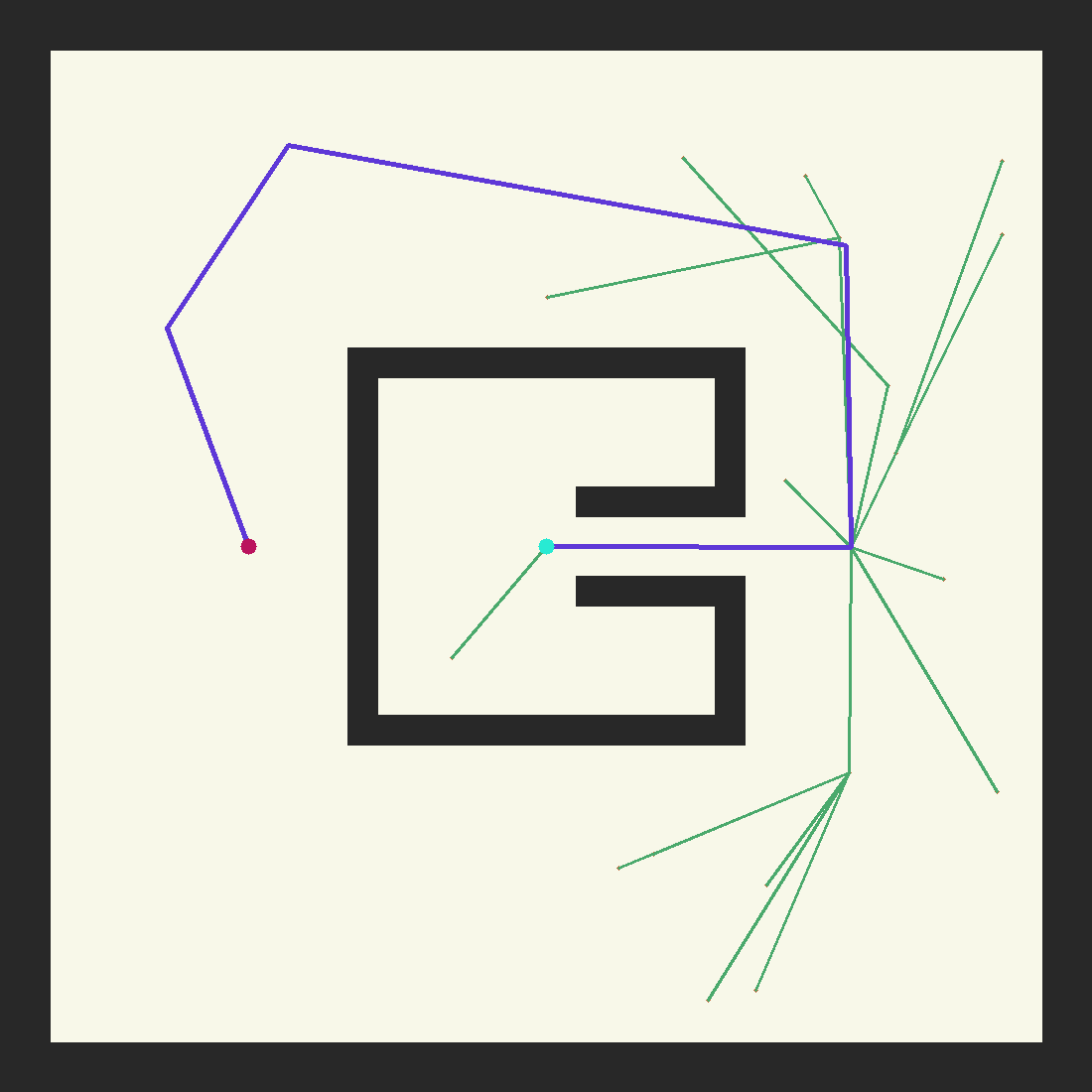}
                \end{minipage}%
            }
            \subfigure[]{
                \begin{minipage}[t]{0.235\linewidth}
                    \centering
                    \includegraphics[width=0.9\linewidth]{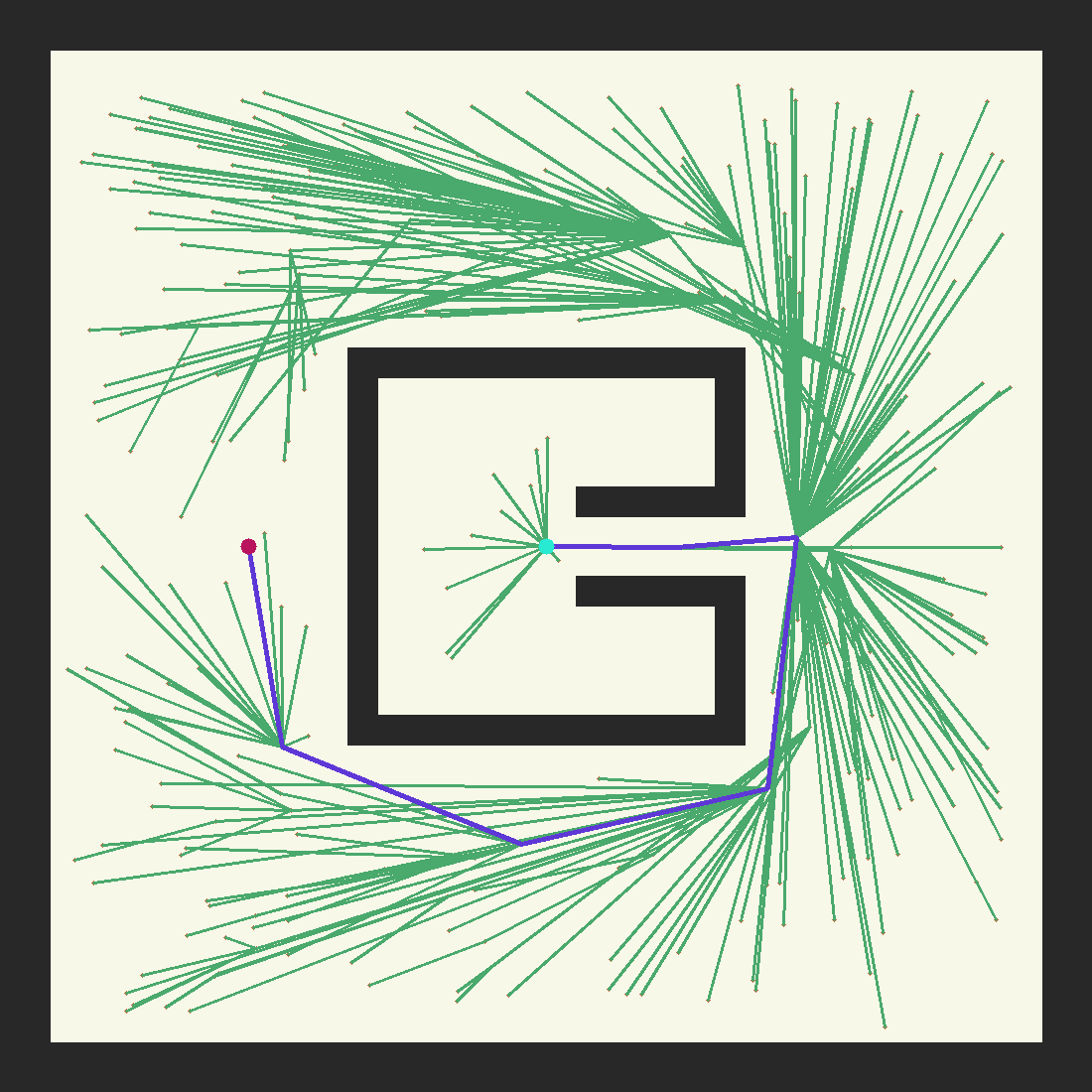}
                \end{minipage}
            }%
            \subfigure[]{
                \begin{minipage}[t]{0.235\linewidth}
                    \centering
                    \includegraphics[width=0.9\linewidth]{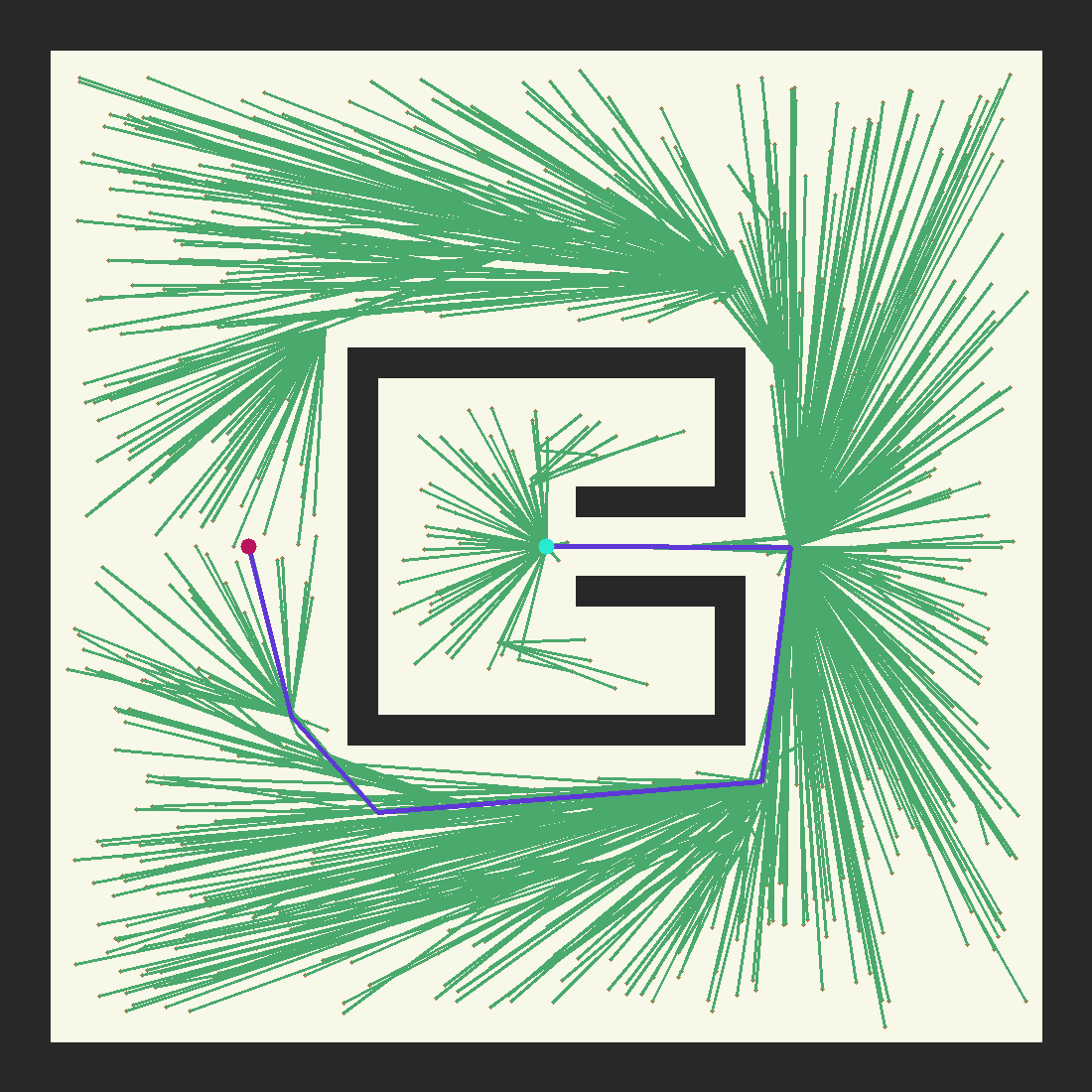}
                \end{minipage}
            }%
        \end{minipage}%
    \centering
    \\
    The RRT\# algorithm.

        \begin{minipage}[t]{1\linewidth}
            \subfigure[]{
                \begin{minipage}[t]{0.235\linewidth}
                    \centering
                    \includegraphics[width=0.9\linewidth]{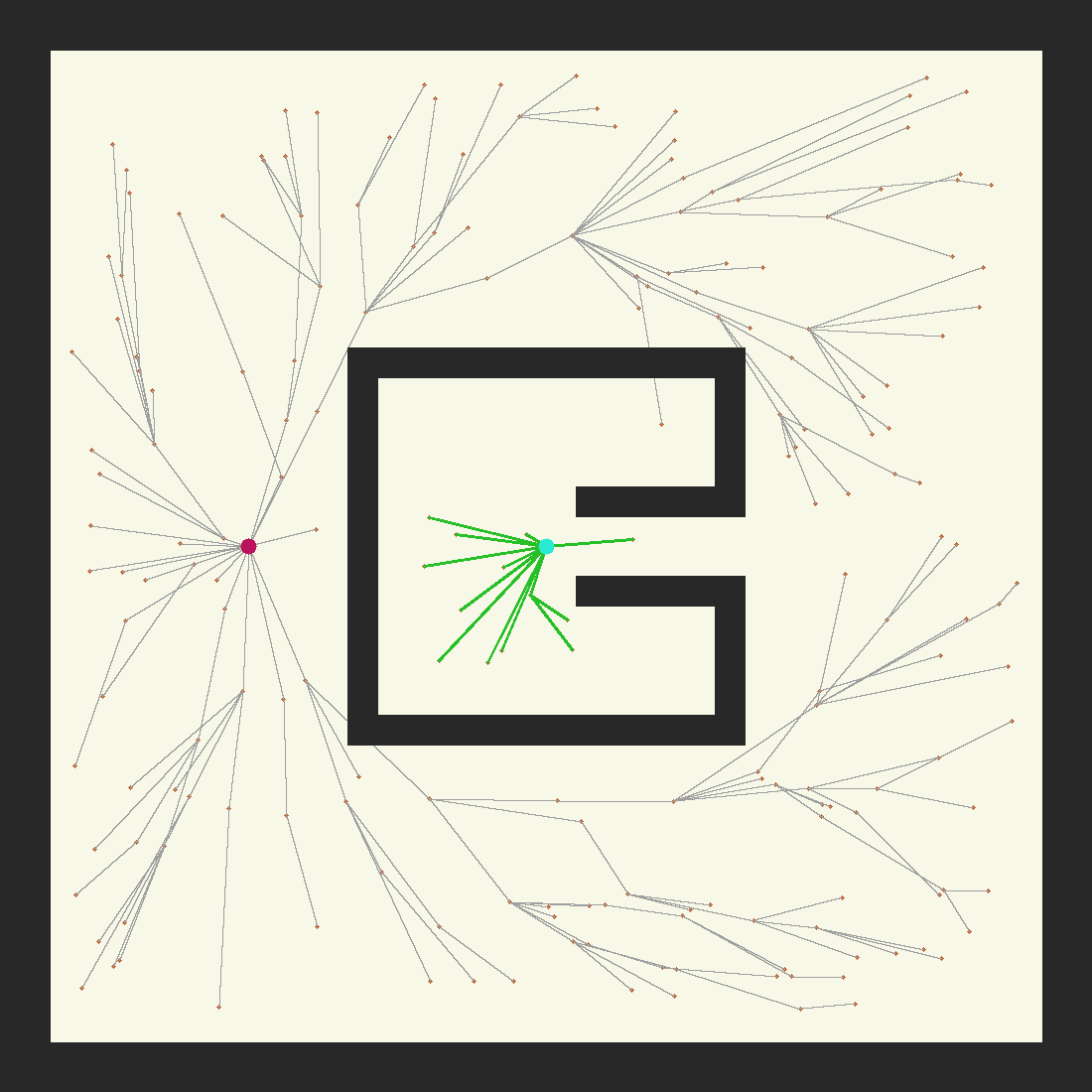}
                \end{minipage}%
            }
            \subfigure[]{
                \begin{minipage}[t]{0.235\linewidth}
                    \centering
                    \includegraphics[width=0.9\linewidth]{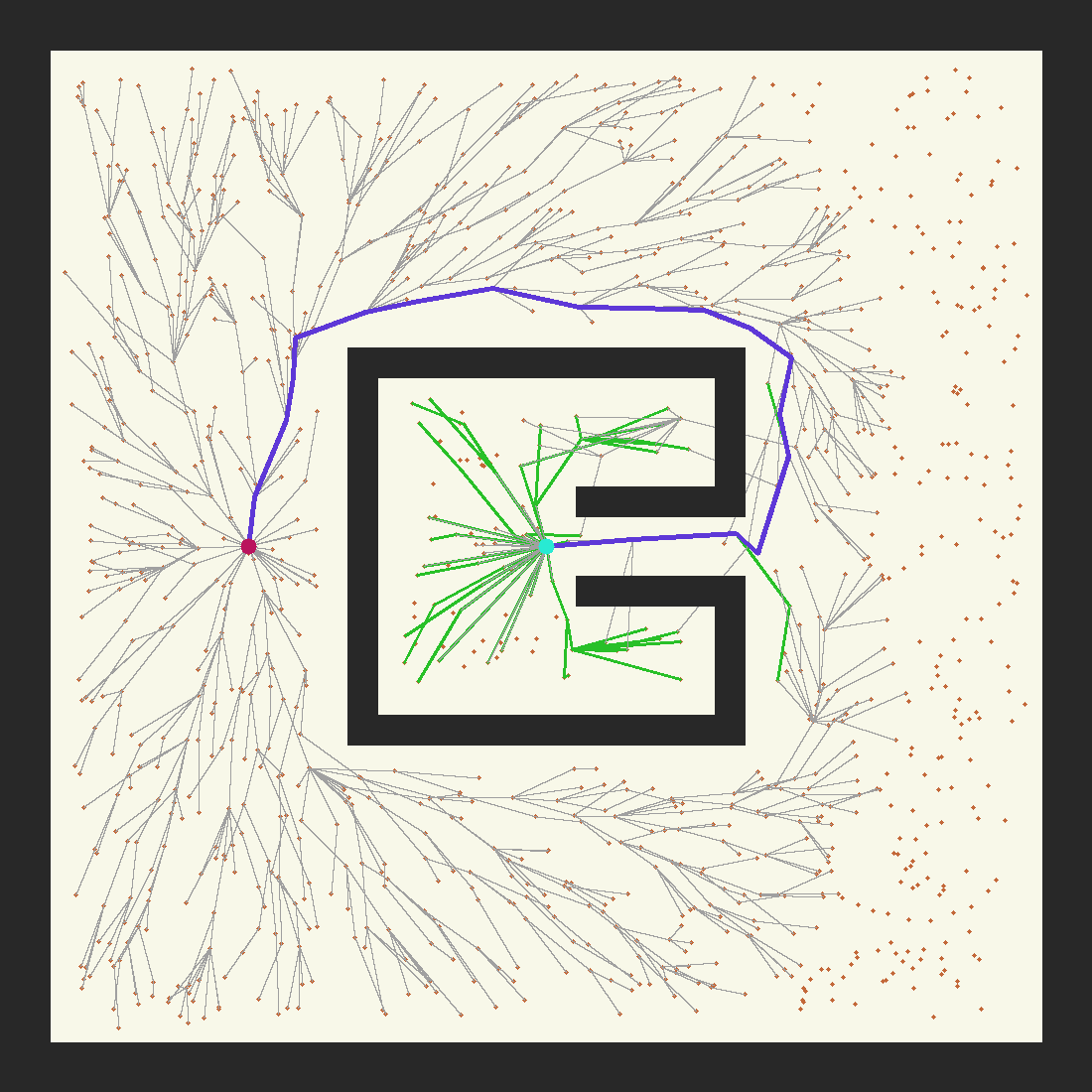}
                \end{minipage}%
            }
            \subfigure[]{
                \begin{minipage}[t]{0.235\linewidth}
                    \centering
                    \includegraphics[width=0.9\linewidth]{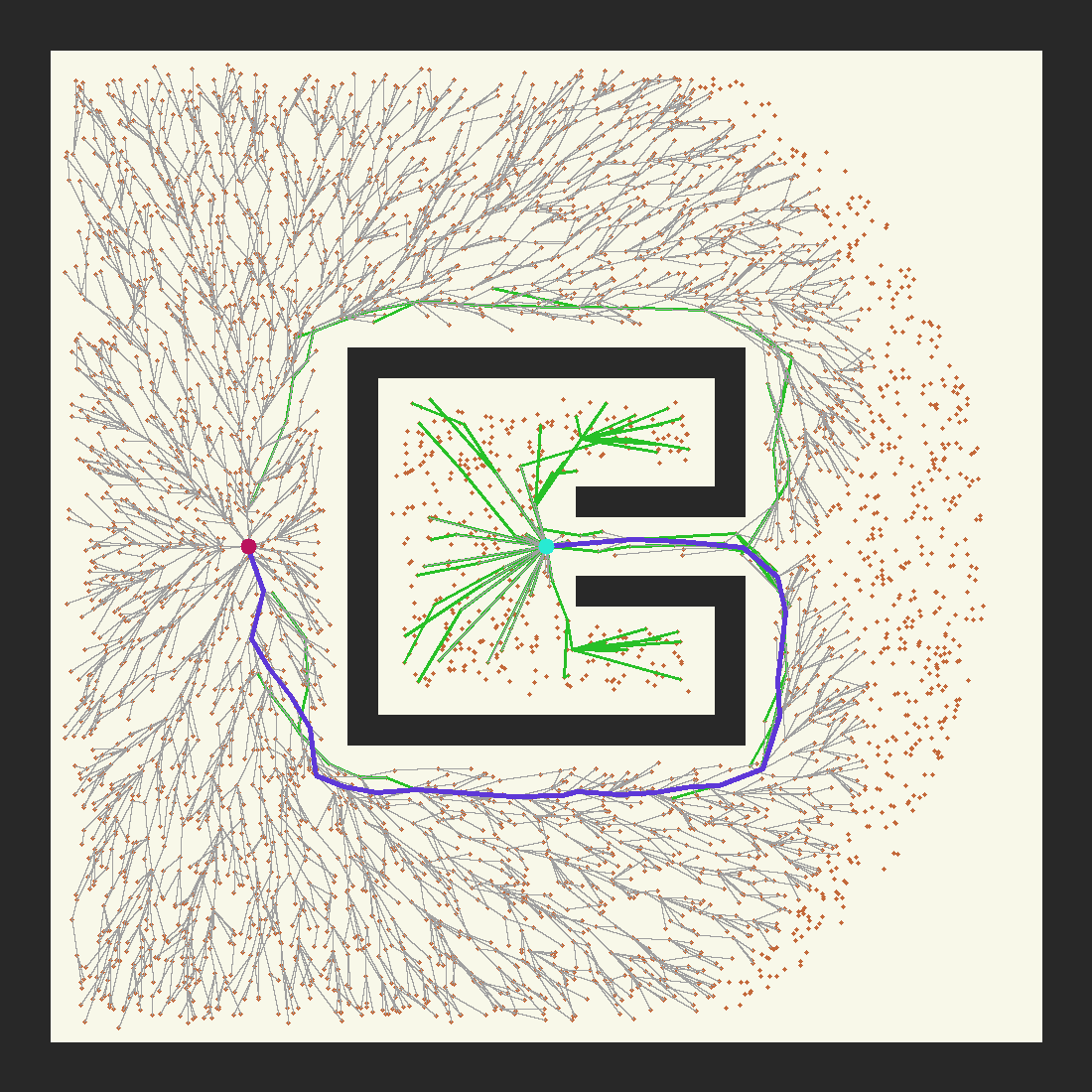}
                \end{minipage}
            }%
            \subfigure[]{
                \begin{minipage}[t]{0.235\linewidth}
                    \centering
                    \includegraphics[width=0.9\linewidth]{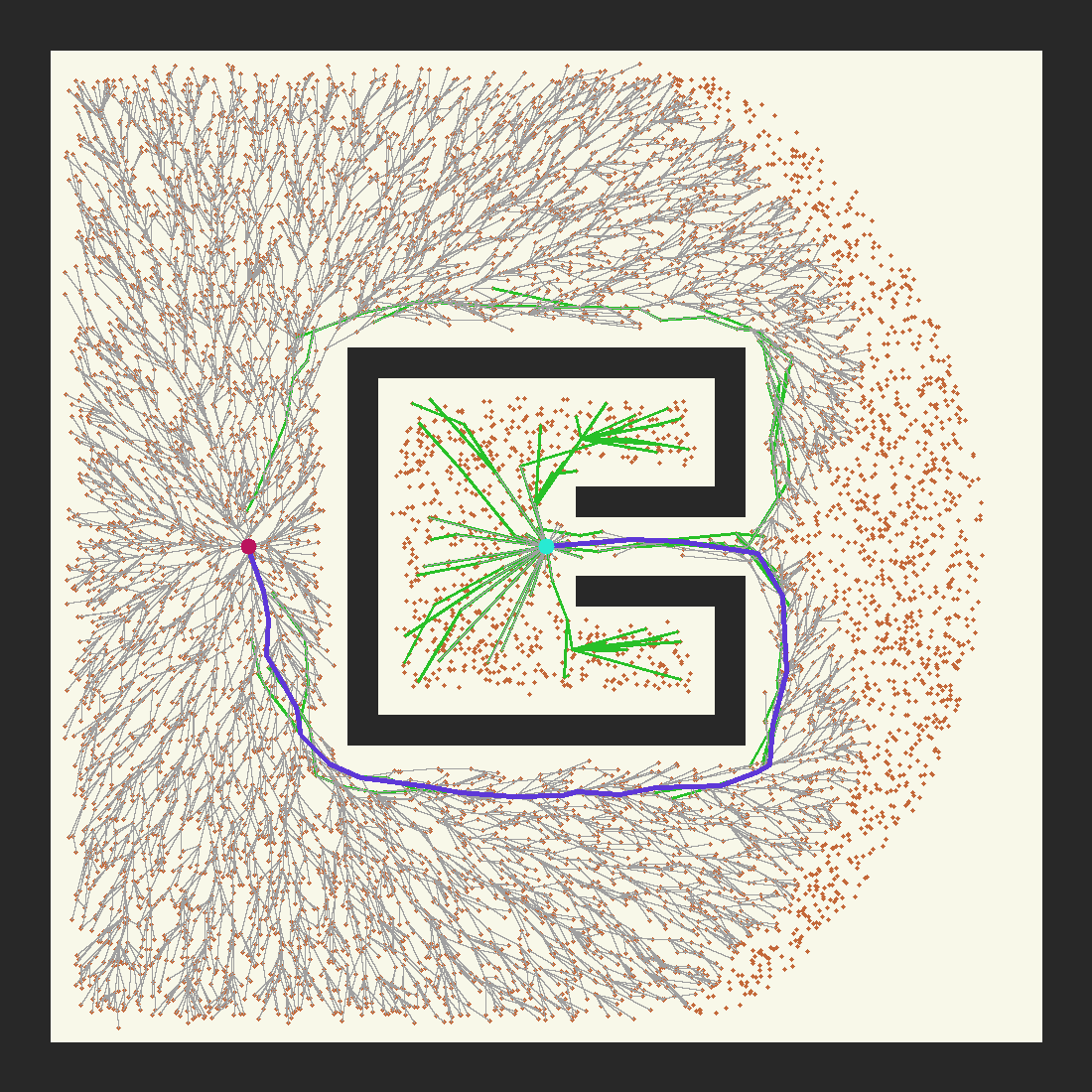}
                \end{minipage}
            }%
        \end{minipage}%
    \centering
    \\
    The AIT* algorithm.

    \subfigure{
        \begin{minipage}[t]{1\linewidth}
            \subfigure[]{
                \begin{minipage}[t]{0.23\linewidth}
                    \centering
                    \includegraphics[width=0.9\linewidth]{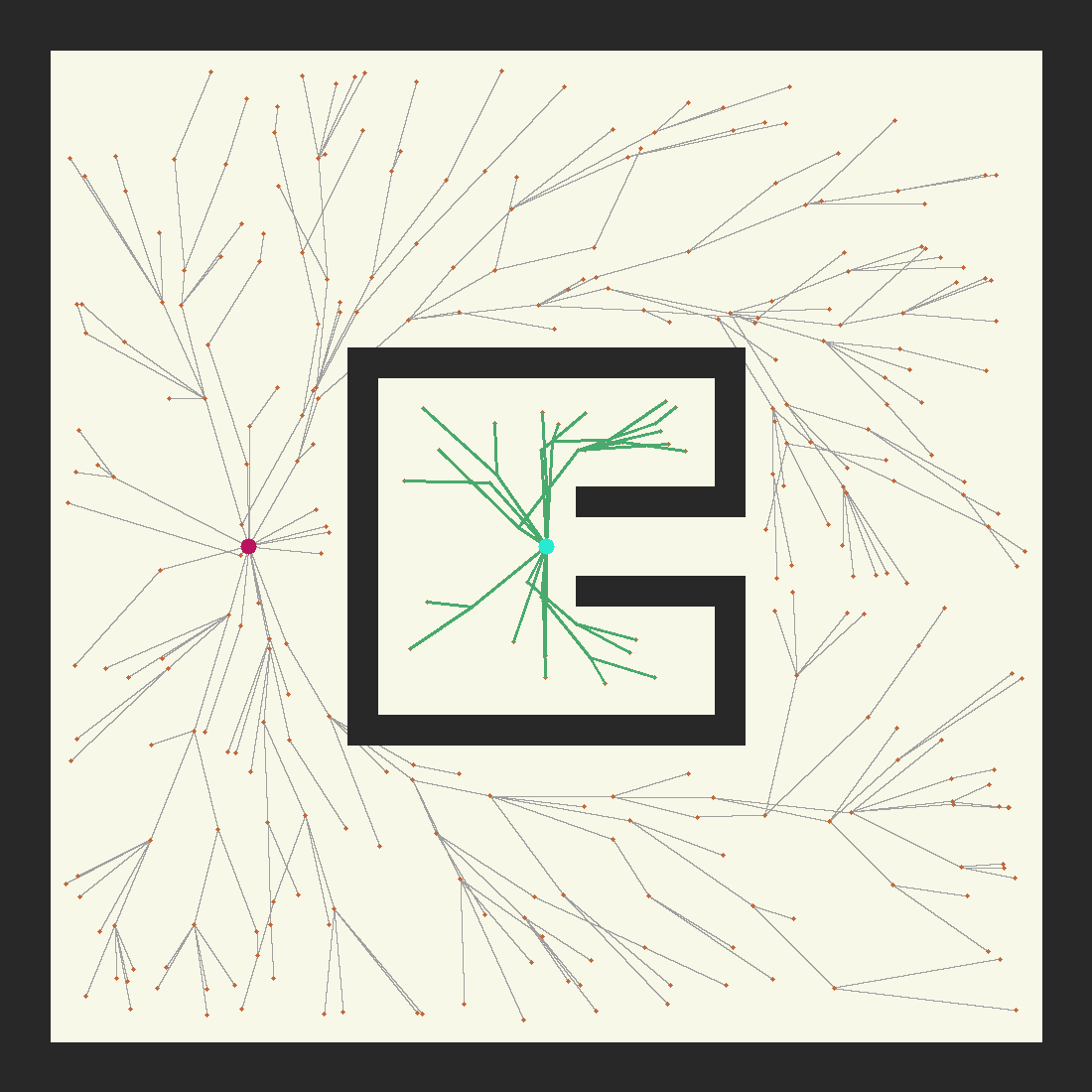}
                \end{minipage}%
            }
            \subfigure[]{
                \begin{minipage}[t]{0.23\linewidth}
                    \centering
                    \includegraphics[width=0.9\linewidth]{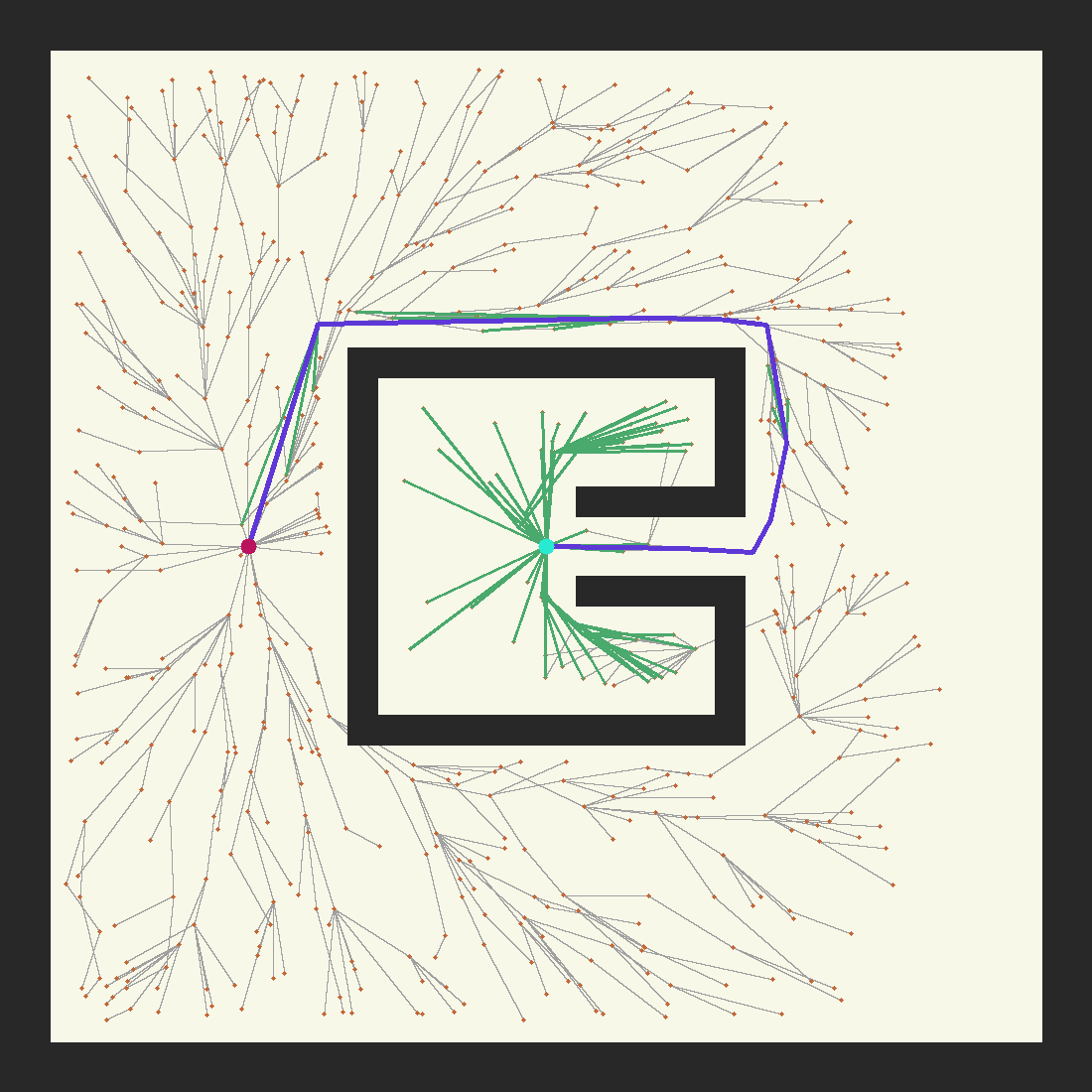}
                \end{minipage}%
            }
            \subfigure[]{
                \begin{minipage}[t]{0.23\linewidth}
                    \centering
                    \includegraphics[width=0.9\linewidth]{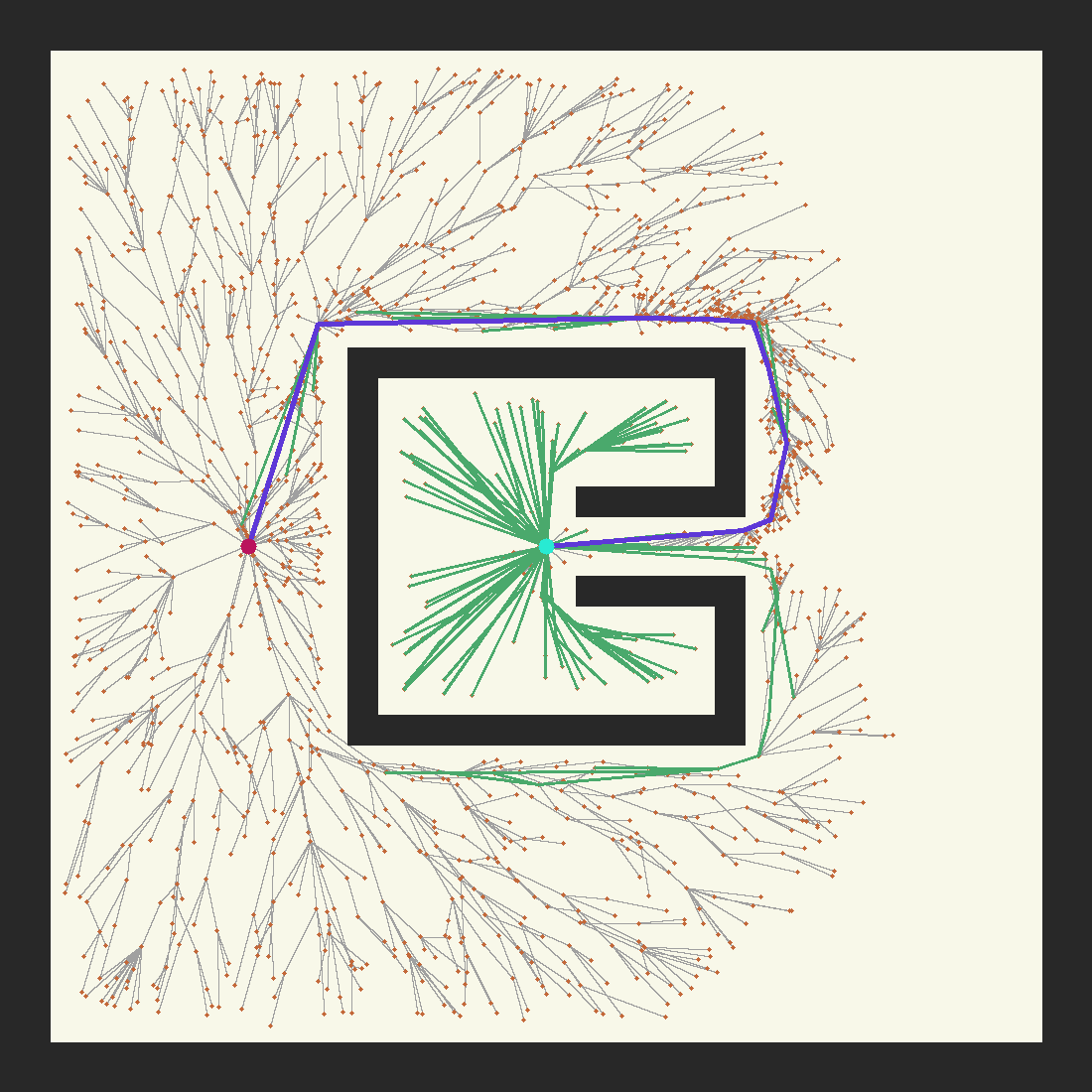}
                \end{minipage}
            }%
            \subfigure[]{
                \begin{minipage}[t]{0.23\linewidth}
                    \centering
                    \includegraphics[width=0.9\linewidth]{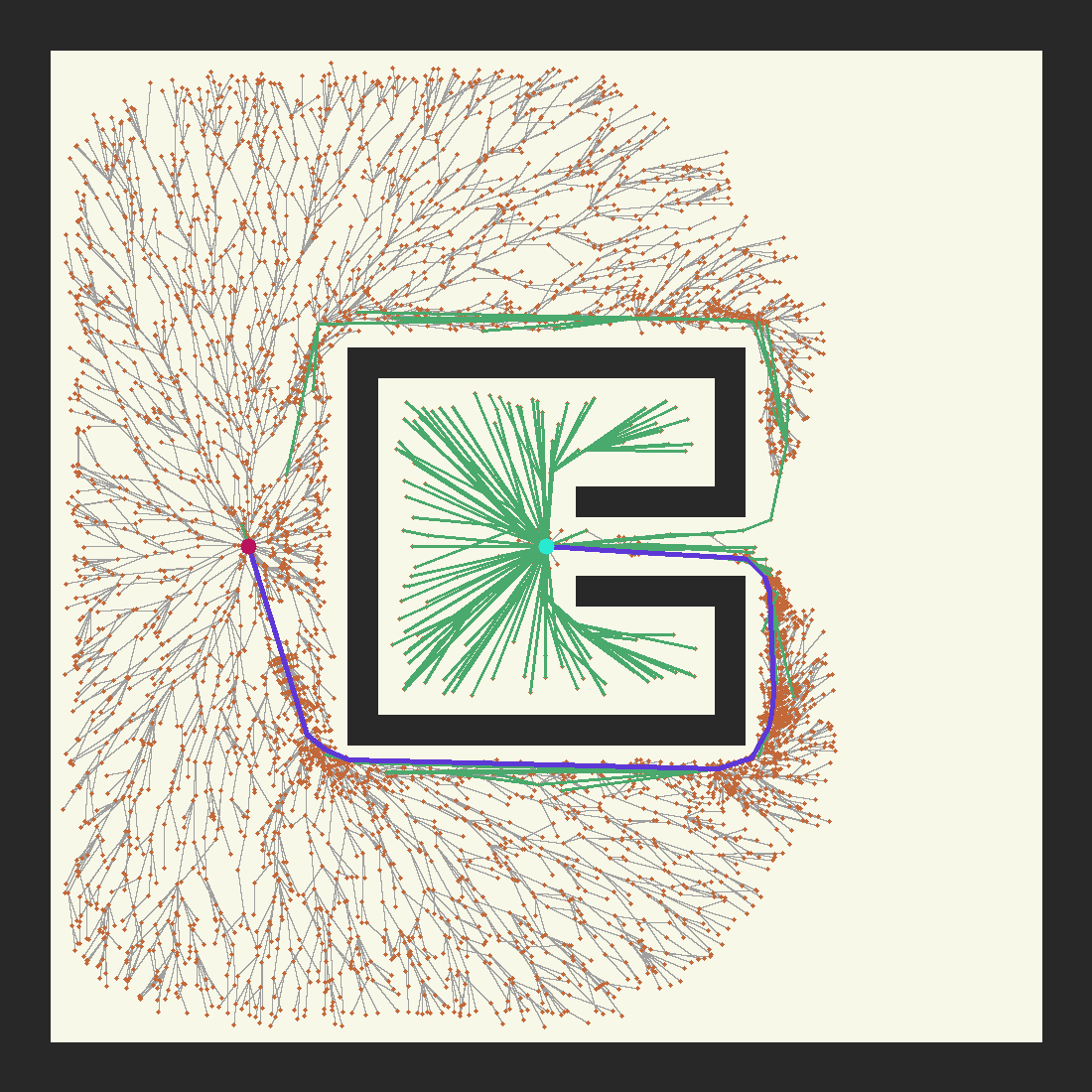}
                \end{minipage}
            }%
        \end{minipage}%
    }
    \centering
    \\
    Our method.

    \centering
    \caption{ Planning procedures of the RRT\# algorithm, the AIT* algorithm, and our method.
              The planning problem is set in the 'BugTrap' environment provided by the OMPL benchmark platform.
              We choose its start state and goal region to make there exists two different optimal solutions, which pass through different zones of the space and have the same solution cost. 
              The optimization objective is set as minimizing the path length.
              It can be seen that our method output a better path than the other two algorithms.
            }
    \label{PlanningProcedure}
\end{figure*}

The planning problem is defined by the state space $\mathcal{X}$, together with the start point $x_{start}$ and goal region $\mathcal{X}_{goal}$. 
The $\mathcal{T}_{\mathcal{F}}$, $\mathcal{T}_{\mathcal{R}}$, and $\mathcal{V}_{sol}$ are initialized as the $\emptyset$.
Our planner will try to find the optimal solution to the problem described in the (\ref{TrajectoryLengthOptimizationObjective}).

In the proposed method, we separate the sampling stage and the searching stage into two explicitly different procedures instead of performing them alternatively in each iteration. 
When processing the sampling stage, we sample a batch of points with the proposed sampling strategy.
In the searching stage, we perform the lazy reverse-searching and construct the $\mathcal{T}_{\mathcal{R}}$ implicitly, then expand the $\mathcal{T}_{\mathcal{F}}$ in terms of the adaptively heuristic estimation provided by the $\mathcal{T}_{\mathcal{R}}$. 
The replanning function will guarantee that all promising vertices in $\mathcal{T}_{\mathcal{F}}$ are optimal under the current topology abstraction. 
In the next iteration, samples in the previous batch are reused as long as they satisfy the inequation described in (\ref{RelevantRegion}).
The $\mathcal{T}_{\mathcal{R}}$ will be disposed of by the end of each iteration, we re-construct the $\mathcal{T}_{\mathcal{R}}$ from the sketch instead of rewiring it in the next iteration, this is due to the concern of the time efficiency.
The procedure of constructing the $\mathcal{T}_{\mathcal{R}}$ can be viewed as a simplified version of the heuristic-based FMT* algorithm without the edge evaluation.
Based on the proposed ideas, we designed the basic workflow of our method, as shown in the Algorithm. \ref{MainFunction}.

In the Line. \ref{pruneFunction} of Algorithm. \ref{MainFunction}, we include the graph pruning method to keep the cardinalities of forward and reverse trees as small as possible.
Graph pruning can enhance query efficiency.
If a sample is not connected to both the $\mathcal{T}_{\mathcal{F}}$ and the $\mathcal{T}_{\mathcal{R}}$ in the previous iteration or the sample is outside the promising region, we will prune it.

The details of the sampling method in Line. \ref{SampleLine} in Algorithm. \ref{MainFunction} is illustrated in Algorithm. \ref{SampleFunction}, the $m$ represents the batch size and the $P_{Inf}$ shows the propability we sample in the Informed space.
The $h_{\mathcal{T}_{\mathcal{F}}}(v)$ and $h_{\mathcal{T}_{\mathcal{R}}}(x)$ always over-estimate the cost, which means the Relevant Region with adaptive heuristic estimation itself can not guarantee the probabilistic completeness (adimissible heuristic in heuristic-based methods). 
Therefore, the samples in our method are generated in mixed mode. 
We use the direct sampling method to sample in the Informed space with the probability $P_{Inf}$. 
The function $\text{WeightVertices}(\mathcal{T}_{\mathcal{F}})$ will weight every vertex $v$ in the $\mathcal{T}_{\mathcal{F}}$ and put the weighted vertices into a priority queue $\mathcal{Q}$. 
The weight $\mathcal{W}$ of a particular vertex is designed according to its selected times $n_{s}$, outgoing edges $n_{o}$, and adaptively heuristic.
The weighting method is shown in (\ref{WeightFunction}), where the $\lambda_1$, $\lambda_2$, and $\lambda_3$ modulate the behavior of our weighting function \cite{joshi2020relevant}.
And the adaptively heuristic is normalized by the current optimal solution.
When generating the new samples, the construction of $\mathcal{T}_{\mathcal{R}}$ upon the current batch is not yet started.
Therefore, the estimated cost-to-go of state in $\mathcal{X}_{\mathcal{S}_{reuse}}$ comes from the $\mathcal{T}_{\mathcal{R}}$ of the last iteration, 
the newly added sample uses the optimal value of $h_{\mathcal{T}_{\mathcal{R}}}(x_{near}) + d(x, x_{near})$, where $h_{\mathcal{T}_{\mathcal{R}}}(x_{near})$ is the cost-to-go of its neighbors.
All samples in $\mathcal{X}_{\mathcal{S}}$ are stored in the GNAT.

\begin{figure*}[t]
    \centering
    \begin{minipage}[t]{1\linewidth}
        \subfigure[]{
            \begin{minipage}[t]{1\linewidth}
                \centering
                \includegraphics[width=1.0\textwidth]{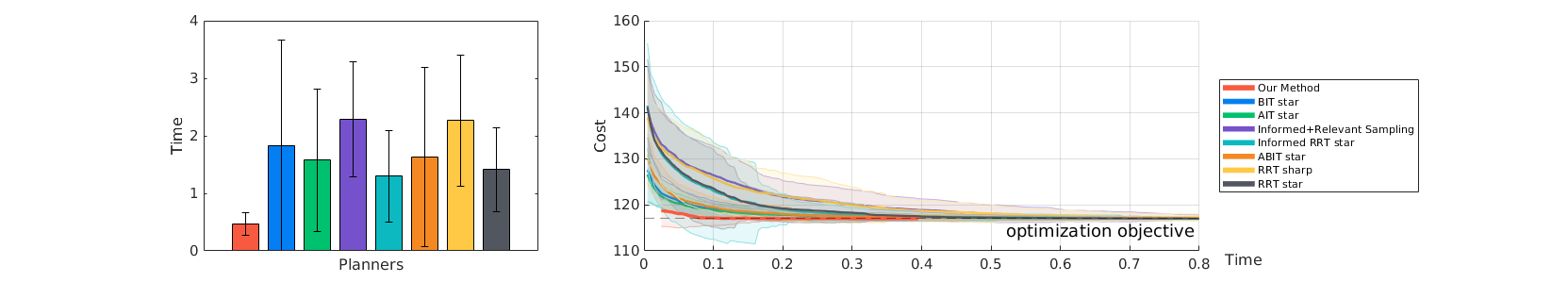}
            \end{minipage}%
        }
    \end{minipage}%

    \centering
    \begin{minipage}[t]{1\linewidth}
        \subfigure[]{
            \begin{minipage}[t]{1\linewidth}
                \centering
                \includegraphics[width=1.0\textwidth]{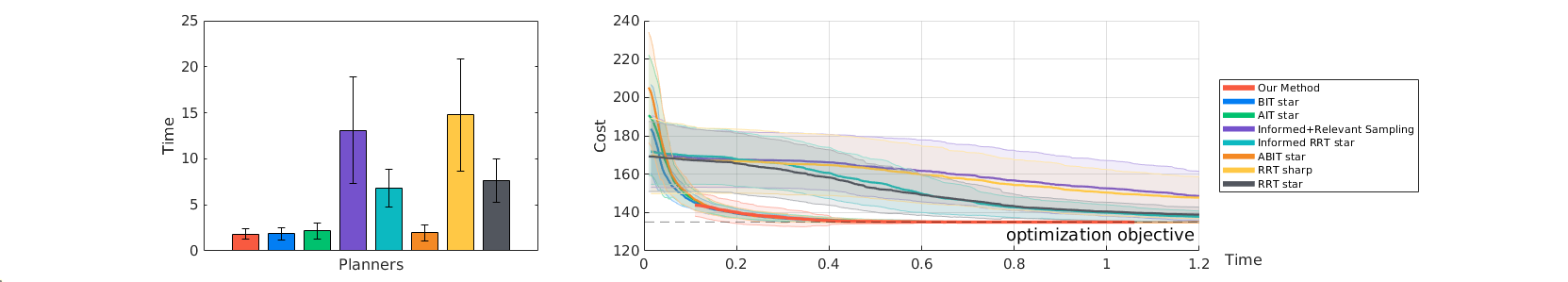}
            \end{minipage}%
        }
    \end{minipage}%

    \centering
    \begin{minipage}[t]{1\linewidth}
        \subfigure[]{
            \begin{minipage}[t]{1\linewidth}
                \centering
                \includegraphics[width=1.0\textwidth]{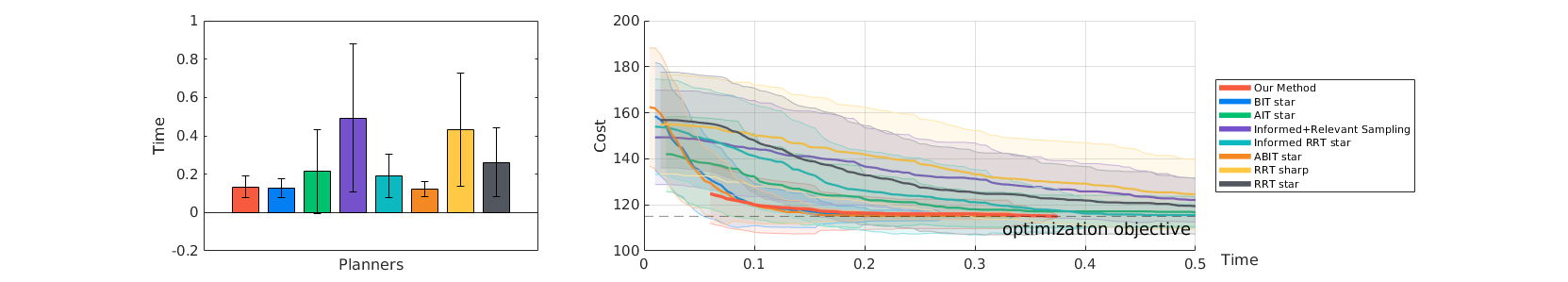}
            \end{minipage}%
        }
    \end{minipage}%
\caption{(a), (b), and (c) show the 2D simulation result in `BugTrap', the `Maze', and the `RandomPolygons' environments, 
    where the left pictures are the time each planner spent to meet the optimization objective 
    and the right pictures are the cost variations over time.
    Planners try to meet the optimization objective, dashed lines in the right pictures show the cost value of the optimization objective.
}
\label{SimulationResults_2D}
\end{figure*}

\begin{figure}[t]
    \centering
    \includegraphics[width=0.46\textwidth]{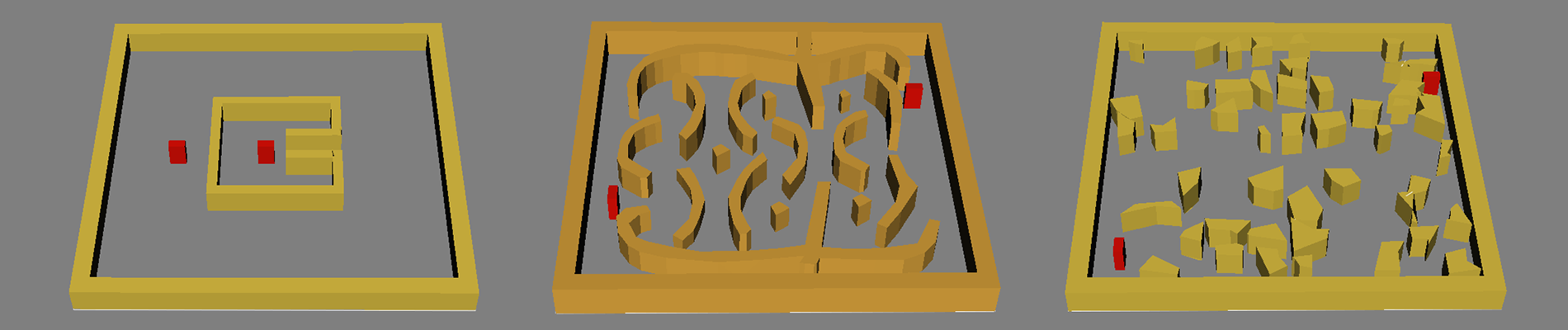}
    \caption{The 2D simulation environments. They are the 'BugTrap', the 'Maze', and the 'RandomPolygons' in the OMPL benchmark platform, respectively. The red cuboid shows the start state and the state in the goal region.}
    \label{SimulationEnvironments_2D}
\end{figure}

\begin{figure}[t]
    \centering
    \includegraphics[width=0.46\textwidth]{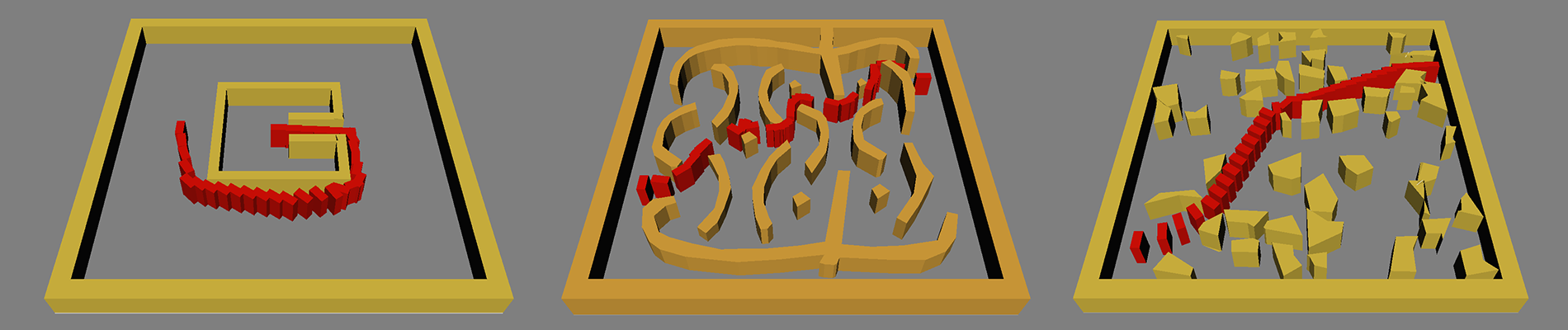}   
    \caption{The trajectories found by our planner in the 'BugTrap', the 'Maze', and the 'RandomPolygons' in the OMPL benchmark platform.}
    \label{SimulationPath_2D}
\end{figure}

The function $SampleRel(Q)$ in Line. \ref{RelevantSampleLine}, Algorithm. \ref{SampleFunction} is the method we use to generate samples with the proposed Relevant Region sampling strategy.
Details are shown in Algorithm. \ref{SampleRelevantRegion}. 
The variable $ExpandTimes$ shows the number of samples generated in terms of vertex $v_{elem}$ with different distortions.
The function $\text{SampleInf}()$ will take $m - \mathcal{X}_{\mathcal{S}_{new}}.size()$ samples in the Informed space \cite{gammell2018informed}.


We utilize the idea of avoiding evaluating every edge in the forward-searching, which comes from the BIT* algorithm \cite{gammell2020batch}.
We illustrate the details of the function $\text{BuildForwardTree}(\mathcal{T}_{\mathcal{F}}, \mathcal{T}_{\mathcal{R}})$ (Line. \ref{MainFunction}, Algorithm. \ref{BuildForwardTreeLine}) in Algorithm. \ref{BuildForwardTreeFunction}.
The vertices in the $\mathcal{T}_{\mathcal{F}}$ will be put in an ordered queue $\mathcal{Q}_{V}$, the $\mathcal{Q}_{V}$ is sorted in terms of $\hat{f}(v) = h_{\mathcal{T}_{\mathcal{F}}}(v) + h_{\mathcal{T}_{\mathcal{R}}}(v)$. 
The $\mathcal{Q}_{E}$ is an ordered queue for the promising edges, which is sorted in terms of $h_{\mathcal{T}_{\mathcal{F}}}(v) + d(v, \hat{x}) + h_{\mathcal{T}_{\mathcal{R}}}(\hat{x})$.
Function $\hat{c}(v_{min}, x_{min})$ and $c(v_{min}, x_{min})$ calculate the estimated cost and the actual cost between vertex $v_{min}$ and state $x_{min}$.
The function $\text{Replan}(\mathcal{T}_{\mathcal{F}})$ guarantees the $\mathcal{T}_{\mathcal{F}}$ is optimal under current space abstraction \cite{arslan2013use}.

Whenever the planner generates a new feasible path connecting the start and the goal, it will trace back along the $\mathcal{T}_{\mathcal{F}}$ and compare it with the current optimal solution path. 
If the newly added path is better, the planner will output a new solution path and use the graph pruning method to restrict the samples to a more compact region.
Otherwise, the planning process will continue until the termination condition is met.

\section{Simulations}

The simulations are all carried through the benchmark platform of the OMPL \cite{sucan2012open} \cite{moll2015benchmarking}.
To validate the generalization ability of our method, we solve the planning problem in both the $SE(2)$ and $SE(3)$ state spaces with our method and several state-of-art algorithms.
In simulation environments, the state spaces are continuous. 
The algorithms that have been tested are all based on random sampling and take samples from these continuous state spaces without discretizing the space.
The robot is represented by a collection of convex polyhedrons and occupies a certain volume.

\subsection{Qualitative Analysis}

\begin{figure*}[t]
    \centering
    \begin{minipage}[t]{1\linewidth}
        \subfigure[]{
            \begin{minipage}[t]{1\linewidth}
                \centering
                \includegraphics[width=1.0\textwidth]{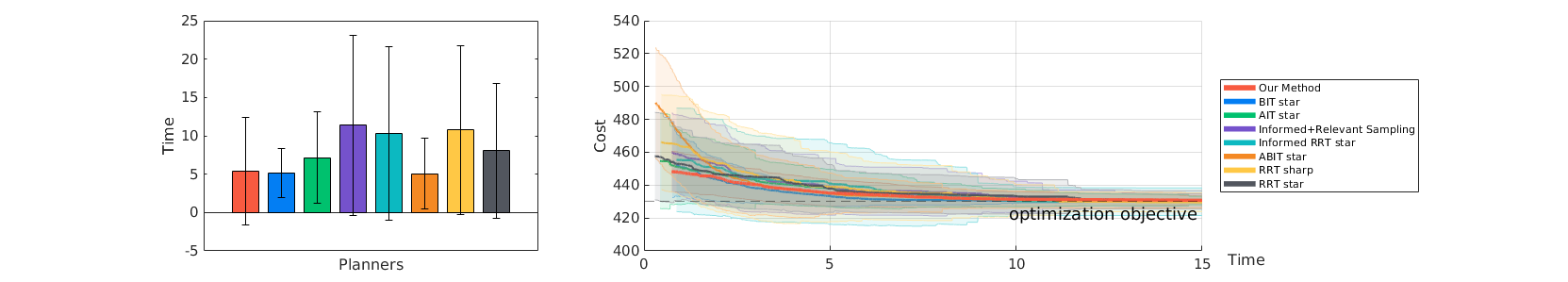}
            \end{minipage}%
        }
    \end{minipage}%

    \centering
    \begin{minipage}[t]{1\linewidth}
        \subfigure[]{
            \begin{minipage}[t]{1\linewidth}
                \centering
                \includegraphics[width=1.0\textwidth]{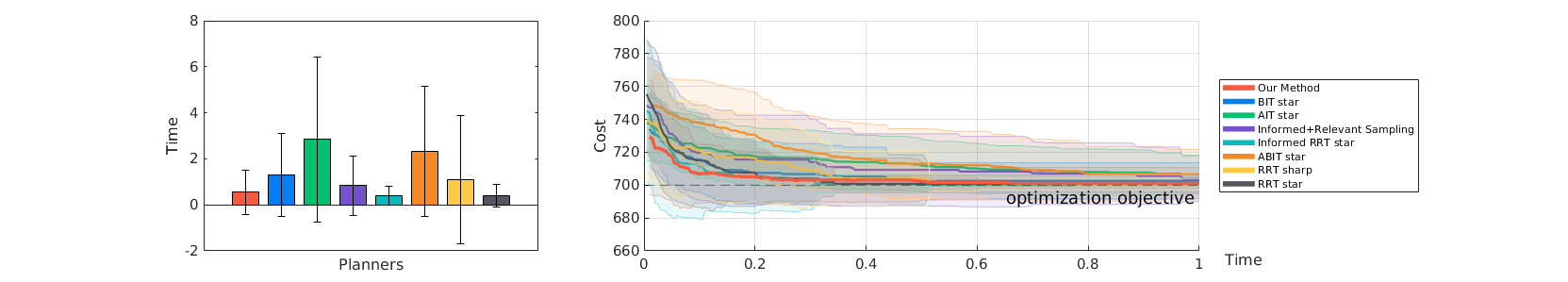}
            \end{minipage}%
        }
    \end{minipage}%
    \caption{(a) and (b) show the 2D simulation result in `BugTrap', the `Maze', and the `RandomPolygons' environments, where the left pictures are the time each planner spent to meet the optimization objective 
    and the right pictures are the cost variations over time.
    Planners try to meet the optimization objective, dashed lines in the right pictures show the cost value of the optimization objective.
    }
\label{SimulationResults_3D}
\end{figure*}

\begin{figure}[t]
    \centering
    \includegraphics[width=0.46\textwidth]{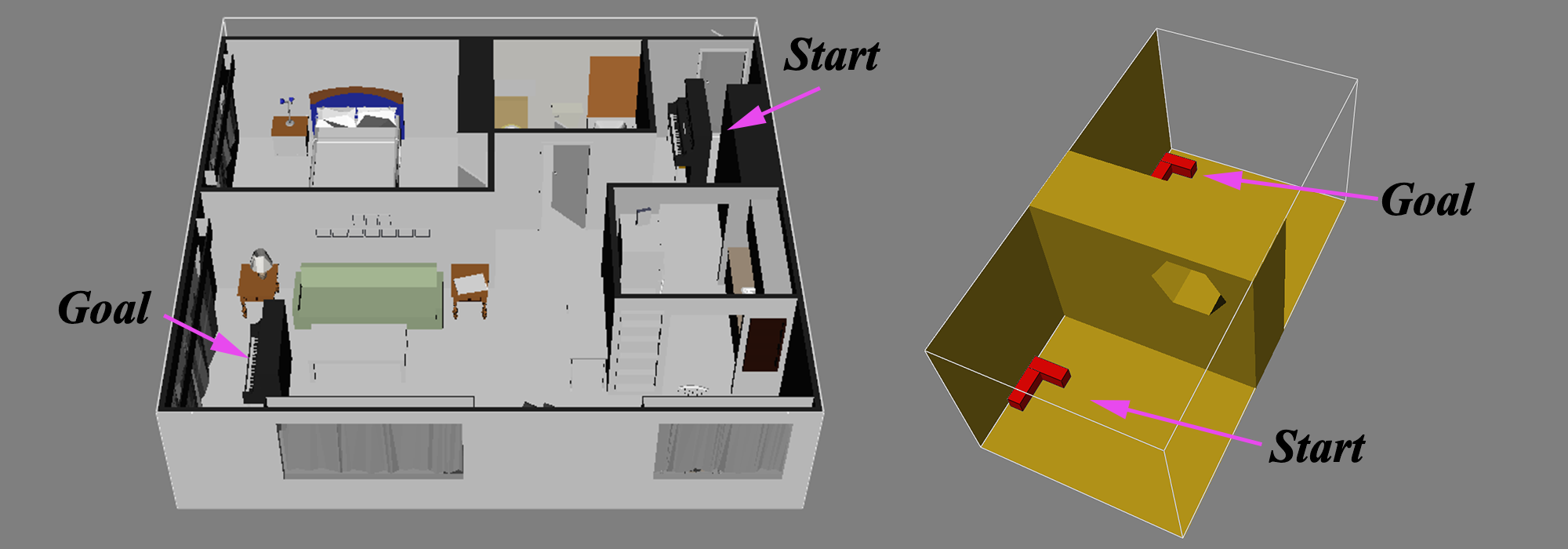}
    \caption{The 3D simulation environments.}
    \label{SimulationEnvironments_3D}
\end{figure}

To give more detail explanation about our approach, we employ the RRT\# \cite{arslan2013use}, AIT* \cite{strub2020adaptively}, and our own method to address the path planning challenge within the `BugTrap' OMPL benchmark environment.
The state space of the `BugTrap' environment is the $SE(2)$ state space, which is composed of the position $x$, $y$ and the orientation $w$.
The planning procedures of the RRT\# \cite{arslan2013use}, the AIT* \cite{strub2020adaptively}, and our method are illustrated in Fig. \ref{PlanningProcedure}, where obstacles, the free space, the start state, the goal region, and vertices are indicated with black, ivory white, pale blue, wine, and orange color, respectively.
We use the dark green lines and violet lines to show the forward tree and the current optimal solution, respectively.
The reverse trees are shown as the grey lines in the figure of the AIT* \cite{strub2020adaptively} and our method.

In the simulation shown in Fig. \ref{PlanningProcedure}, the planning problem contains two optimal solutions, one is to pass through the region upper the obstacle, and the other one is to pass through the lower part.
Fig. \ref{PlanningProcedure} shows that all the methods in Fig. \ref{PlanningProcedure} can acquire the global asymptotical optimality.
Both the AIT* and our method use the lazy reverse-searching tree to guide the sampling and have the graph pruning method to constraint the samples and the trees.
From the (j)-(m) in Fig. \ref{PlanningProcedure}, it can find that both the forward and reverse trees of our method are optimal under current state space abstraction.
In addition, our method concentrates on taking samples in the region with a higher potential to improve the current solution, which can be seen in (m) of Fig. \ref{PlanningProcedure}, our method pays more attention to the turning corners with our direct sampling method.

\subsection{Simulations in $SE(2)$ State Space}

We choose the $SE(2)$ environments shown in Fig. \ref{SimulationEnvironments_2D} to verify our method, and they are called the `BugTrap', the `Maze', and the `RandomPolygons' in the OMPL benchmark platform. 
To give the reader an intuitional understanding of our 2D planning simulations, we show the trajectories found by our method in Fig. \ref{SimulationPath_2D}.
The trajectories are interpolated in terms of time.

In our 2D simulations, the state space definition contains the position $x, y$ and orientation $w$.
To manifest the superiority of our method, we compared with seven different state-of-art algorithms, they are the RRT*, the BIT*, the AIT*, the ABIT*, the Informed RRT*, the RRT\#, and the Informed $+$ Relevant sampling method proposed in \cite{joshi2020relevant}.
In these simulations, we use the trajectory length as the cost metric.
The optimization objective is set as $\beta \times c_{opt}$, where the $c_{opt}$ is the cost of the optimal solution and $\beta$ is a number close to $100\%$.
The $c_{opt}$ is the solution cost of the RRT* method after $300$ seconds of execution, which is nearly optimal. 
We choose to use this number to represent the optimal cost.
To reduce the randomness, each planner runs $100$ times in each environment.

The simulation results in the $SE(2)$ state spaces are shown in Fig. \ref{SimulationResults_2D}.
On the left side of the pictures, the charts shows the amount of time each planner took to generate the required path.
On the right side, the cost distribution is presented in terms of time. We begin plotting the line charts once 50\% of all runs have found a solution, and stop once 95\% have completed the problem-solving process. 
Hence, the speed of obtaining the initial solution can also be displayed in the same chart.
Additionally, error bars are provided for all bar charts and line charts.

The 2D simulation results show that our method acquired significant improvements and achieved better performance.
Both the initial solution quality and the convergence rate of our method are better than the others.
The only drawback of our method is we generate the initial solution slower than the others, but we acquired the best initial solution quality.
And our initial solution is better than the others' optimized solutions at the same time point.

\subsection{Simulations in $SE(3)$ State Space}

Besides the 2D simulation introduced previously, we also carried on the simulation in the $SE(3)$ state space. 
In the 3D simulations, we include the `3D\_Apartment' planning problem from the OMPL benchmark platform \cite{moll2015benchmarking}, which is a `piano movers' problem, as the left environment in Fig. \ref{SimulationEnvironments_3D} shows.
The other simulation is set as a planning problem in 3D narrow passage environment.
In the 3D simulation, planners and their parameter sets are the same as the planners we choose in 2D simulations.
Each planner will solve each planning problem 100 times to avoid the randomness.
The 3D simulation results are shown in Fig. \ref{SimulationResults_3D}.

\section{Conclusions}
In this paper, we suggest improving planning efficiency with a novel sampling strategy. This strategy relies on the optimal cost-to-come and problem-specific cost-to-go heuristics, acquired through the global replanning function and the lazy-reverse searching method, respectively. This sampling strategy generates samples in the Relevant Region to guide the planner to search the most promising regions, improving the quality of the initial solution path and the convergence rate. Simulation results in $SE(2)$ and $SE(3)$ spaces demonstrate that the proposed method achieves better performance when solving the planning problem. In the future, we aim to take advantage of the reverse searching method to further accelerate planning efficiency.


\printcredits

\bibliographystyle{cas-model2-names}

\bibliography{references.bib}

\balance

\end{document}